\documentclass{article}

\usepackage[margin=1in]{geometry}
\usepackage[square,numbers]{natbib}
\usepackage{graphicx}
\usepackage{hyperref}
\usepackage{amsmath}
\usepackage{amssymb}
\usepackage{subcaption}
\usepackage{xargs}
\usepackage{multirow}
\usepackage{sidecap, caption}
\usepackage{booktabs}
\usepackage{comment}
\usepackage{xcolor}

\newcommand\blfootnote[1]{%
  \begingroup
  \renewcommand\thefootnote{}\footnote{#1}%
  \addtocounter{footnote}{-1}%
  \endgroup
}

\newcommand{\real}[1]{\mathbb{R}^{#1}}

\newcommandx{\dd}[3][3={}]{\frac{\text{d}^{#3}{#1}}{\text{d} {#2}^{#3}}}
\newcommandx{\pp}[3][3={}]{\frac{\partial^{#3}{#1}}{\partial {#2}^{#3}}}

\newcommand{\bmat}[1]{\left[\begin{array}{#1}}
\newcommand{\emat}{\end{array}\right]}
\newcommand{\bgrid}[1]{\begin{array}{#1}}
\newcommand{\egrid}{\end{array}}

\begin{document}
	
	\title{Data-driven stochastic reduced-order modeling of \\  parametrized dynamical systems}

	\author{{Andrew F. Ilersich, Kevin Course, Prasanth B. Nair}\blfootnote{\texttt{andrew.ilersich@mail.utoronto.ca}, \texttt{kevin.course@mail.utoronto.ca}, \texttt{prasanth.nair@utoronto.ca}} \\ 
	{\small \em University of Toronto Institute for Aerospace Studies}, \\ {\small \em 4925 Dufferin Street, Toronto, ON M3H 5T6}}
	
	\date{} 

	\maketitle
	
	\begin{abstract}
		Modeling complex dynamical systems under varying conditions is computationally intensive, often rendering high-fidelity simulations intractable. Although reduced-order models (ROMs) offer a promising solution, current methods often struggle with stochastic dynamics and fail to quantify prediction uncertainty, limiting their utility in robust decision-making contexts. To address these challenges, we introduce a data-driven framework for learning continuous-time stochastic ROMs that generalize across parameter spaces and forcing conditions. Our approach, based on amortized stochastic variational inference, leverages a reparametrization trick for Markov Gaussian processes to eliminate the need for computationally expensive forward solvers during training. This enables us to jointly learn a probabilistic autoencoder and stochastic differential equations governing the latent dynamics, at a computational cost that is independent of the dataset size and system stiffness. Additionally, our approach offers the flexibility of incorporating physics-informed priors if available. Numerical studies are presented for three challenging test problems, where we demonstrate excellent generalization to unseen parameter combinations and forcings, and significant efficiency gains compared to existing approaches.
	\end{abstract}
	
	\section{Introduction}
	
	Numerical simulation of complex dynamical systems is computationally demanding, particularly in applications requiring multiple simulation runs under varying parameter settings and forcing conditions such as optimal design, parameter estimation, and control. 
	Model order reduction is a powerful approach to address this challenge, wherein a reduced-order model (ROM) over a lower-dimensional latent space is constructed to serve as a computationally efficient surrogate/emulator of the full-order model (FOM). Arguably, the earliest work on this topic dates back to Fox and Miura in 1971~\cite{FoxMiura1971}, who proposed projection schemes for  accelerating structural design calculations. There is now a rich body of literature on the theoretical and computational aspects of projection-based ROMs for a broad class of parametrized operator problems~\cite{Benner2022}. 
	
	A major limitation of projection-based ROM approaches is that they require direct access to the FOM.  In many practical applications, the underlying FOM may be inaccessible, too complex to work with, or entirely unknown with the trajectory observations arising from experimental studies. This has motivated the development of data-driven ROM algorithms capable of learning directly from observed FOM or experimental trajectories, circumventing the need for explicit knowledge of the governing equations.
	
	Most of the existing ROM approaches, whether projection-based or data-driven, proceed in two steps: first constructing forward and approximate inverse mappings between the full-order state and a low-dimensional latent state, then formulating equations governing the latent dynamics.
	The effectiveness of dimensionality reduction in this first step is fundamentally linked to the Kolmogorov $n$-width of the solution manifold, which quantifies the best possible approximation of the dynamics in a low-dimensional subspace~\cite{Peherstorfer2022}. For problems with rapidly  decaying Kolmogorov $n$-width, linear dimensionality reduction techniques such as proper orthogonal decomposition (POD) have proven highly effective~\cite{Audouze2009, Audouze2013, Benner2022,Peherstorfer_2016_Operator_Inference,McQuarrie2023,Wan_2018_ROM}. However, many complex systems exhibit solution manifolds with slowly decaying $n$-width, motivating the use of nonlinear dimensionality reduction approaches~\cite{Xing2015, Xing2016,XuDuraisamy2020, LeeCarlberg2020, Kim_2022_PINN_ROM,Maulik2021, Barnett2022, Romor2023}, basis adaptation algorithms~\cite{Huang_2023_ROM_Chaotic_Problems}, and gradient-based reduction methods~\cite{Baptista_2022_Gradient_Reduction}. 
	Projection-based methods leverage the FOM in the second step~\cite{Benner2015,Benner2022,LeeCarlberg2020,Kim_2022_PINN_ROM}, whereas data-driven approaches employ a non-intrusive approach involving the minimization of an appropriate loss function over training trajectories to estimate the ROM parameters~\cite{Audouze2013, Xing2016, Peherstorfer_2016_Operator_Inference, McQuarrie2023,Fries_2022_LaSDI,DuanHesthaven2024}.
	
	Even though the two-step approach has been applied successfully in many applications, it is worth noting that for certain datasets and latent mappings, the latent state trajectories learned in the first step can cross each other. In such scenarios, it is impossible to represent the dynamics with a system of differential equations~\cite{Dupont_2019_ANODE}. Some studies approach ROM by simultaneously learning the state space mappings and the ROM dynamics, which avoids this issue~\cite{Lee_2021_Parameterized_Neural_ODE_ROM,Champion_2019_Data_Driven_Coordinates, Aka_2025_Balanced_Neural_ODEs}.
	
	Champion et al.~\cite{Champion_2019_Data_Driven_Coordinates} proposed a data-driven ROM method that simultaneously learns the state space mapping along with a sparse identification of nonlinear dynamics (SINDy) ROM. Similar to the approach presented in this work, the SINDy method also does not require a forward solver during training. However, this is achieved by assuming that the time-derivative of the state is available or can be computed using numerical differentiation. In the latter case, the accuracy of the ROM can deteriorate if the measurement noise variance is high. In contrast, the  proposed approach requires only the state measurements to infer a ROM.

	In the present work, we develop an amortized stochastic variational inference (SVI) scheme for learning ROMs across parameter spaces and forcing conditions using only FOM trajectory data. Our end-to-end methodology simultaneously infers a variational autoencoder for state space mapping and a system of continuous-time stochastic differential equations (SDEs) governing the latent dynamics. The present work advances data-driven ROM in the following ways: (1)~by learning SDE-based ROMs, our approach can address challenging problems with stochastic dynamics (see, for example,~\cite{Karniadakis_2005_Microflows, Gillespie_2007_Chemical_Kinetics, Rao_2002_Intracellular_Noise}), (2)~the end-to-end learning approach developed here circumvents the technical challenges associated with two-step methods described earlier, and (3)~by adopting a stochastic modeling and inference framework, our approach enables quantification of prediction uncertainty, which is critical for downstream decision-making applications such as optimal design and control.

	Although learning SDEs in a variational inference setting offers several theoretical advantages, its implementation poses significant computational challenges. Traditional approaches require integrating the latent SDE using a forward solver for each training trajectory to calculate the evidence lower bound and its gradients~\cite{Archambeau_2007_GP_Approx_SDE,Archambeau_2007_Variational_Inference_Diffusion,Li_2020_Scalable_Gradients_VI_SDE}. The need for a forward solver in the training loop, which is a well-known known bottleneck in training neural differential equations in general~\cite{Kidger_2022_Neural_Differential_Equations}, can make the training cost prohibitive for high-dimensional states and large-scale datasets. Moreover, parameter updates during the inference process tends to increase the stiffness of the underlying differential equations, precipitating a blow-up in the cost per iteration. 
	
	We address these challenges by leveraging a reparametrization trick recently proposed by Course and Nair~\cite{Course_2023_State_Estimation} that allows us to train the ROM \emph{without} integrating it, which not only reduces the training cost, but decouples it from the SDE stiffness. Combined with an amortization strategy~\cite{Course_2023_Amortized_Reparameterization}, our approach enables further reductions in training cost by defining variational distributions over short partitions of the time-domain. We present numerical studies on three test cases to demonstrate the efficacy of the proposed ROM approach in comparison to existing approaches. In the final test case, we consider ROM of a controlled fluid flow problem with $105,600$ states, to demonstrate the scalability of our approach.
	
	\section{Methodology}
	
	\begin{figure}[t]
		\centering
		\includegraphics[width=\textwidth]{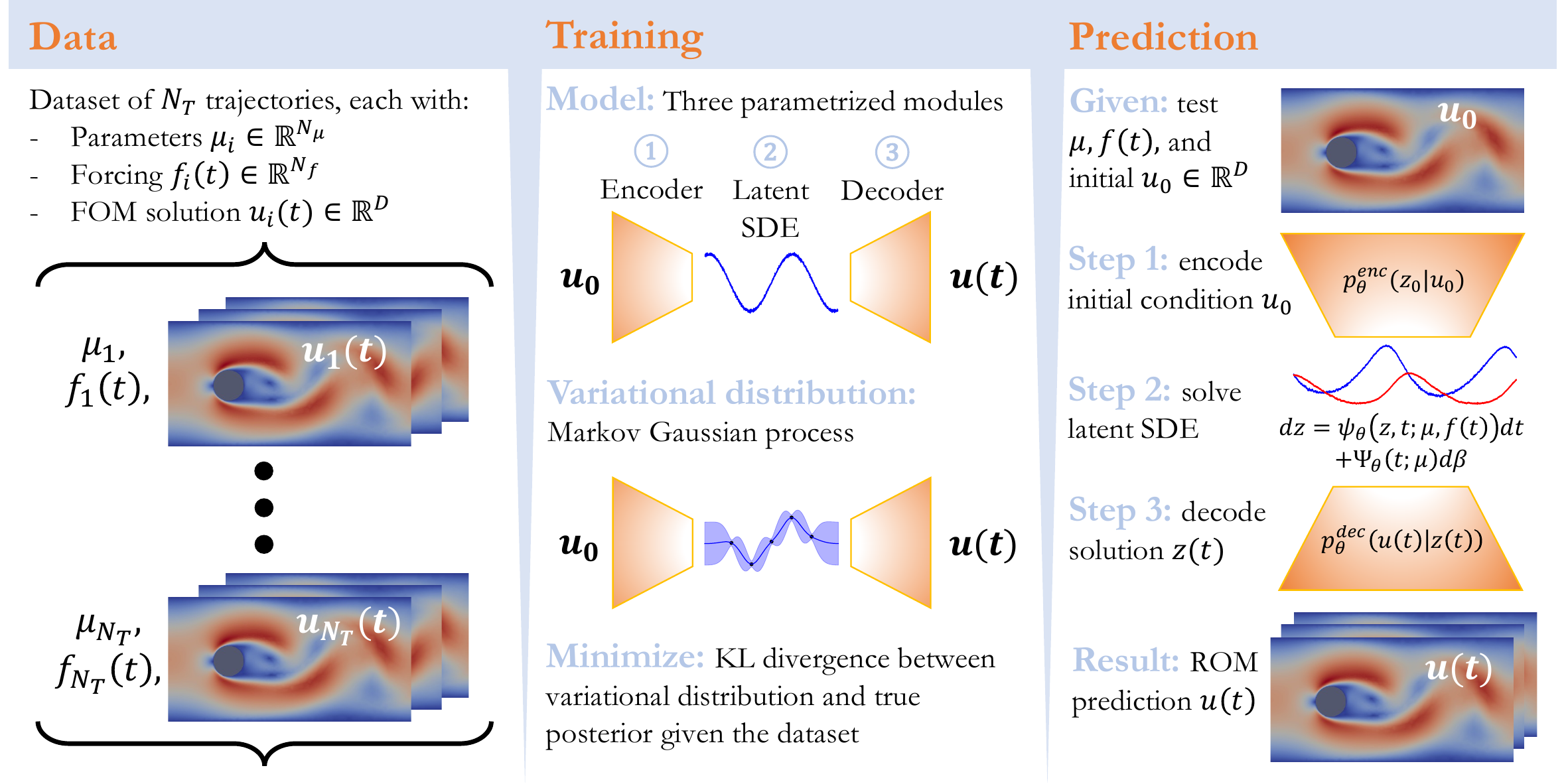}
		\caption{Graphical overview of the proposed stochastic ROM approach. The ROM, which we train by stochastic variational inference, consists of three modules: a probabilistic encoder, a latent SDE, and a probabilistic decoder. Making a prediction with the ROM starts with parameters $\mu$, forcing $f(t)$, and initial condition $u_0$. We encode the initial condition to obtain the corresponding latent $z_0$. We then solve the latent SDE to obtain realizations of the trajectory $z(t)$. Decoding the sampled trajectories yields the stochastic ROM prediction of the FOM QoI, $u(t)$, that can be postprocessed for the mean prediction and statistical error bars.}
		\label{fig:method}
	\end{figure}
	
	\subsection{Problem statement}
	
	Consider a parametrized spatio-temporal quantity of interest (QoI): $\widetilde{u}: \Omega \times [0,T] \times \mathcal{F}_{\mu} \times  \mathcal{F}_f \to \mathbb{R}$, either computed using a FOM or obtained through experimental measurements over the time interval $[0,T]$. Here, $\Omega \subset \mathbb{R}^m$ denotes the spatial domain, while $\mathcal{F}_{\mu} \subset \mathbb{R}^{N_\mu}$ and $\mathcal{F}_f \subseteq L^2(\Omega \times [0,T]; \mathbb{R})$ represent the parameter space and the space of forcing functions, respectively. Our training dataset comprises $N_T$ noisy trajectories of the QoI under varying parameters and forcing functions, i.e., $\mathcal{D}:= \{  ( \mu_i, \widetilde{f}_i, \widetilde{u}_i ) \}_{i=1}^{N_T}$, where $\mu_i \in \mathcal{F}_{\mu}$, $\widetilde{f}_i \in \mathcal{F}_f$, and $\widetilde{u}_i \equiv \widetilde{u}(x,t;\mu_i, \widetilde{f}_i)$ with $x \in \Omega$, and $t \in [0,T]$.
	
	In practice, we work with discrete representations of the QoI and forcing function involving $D$ and $N_f$ terms, respectively, that are obtained through spatial discretization on a mesh or by projecting onto appropriate sets of spatial basis functions. These discrete representations are sampled at time stamps $t_{i,j} \in [0,T]$, where $i=1,2,\ldots,N_T$ indexes trajectories and $j=1,2,\ldots,N_i$ indexes observations within each trajectory. We denote the spatially discretized forcing function and QoI as $f(t) \in \mathbb{R}^{N_f}$ and $u(t; \mu, f) \in \mathbb{R}^D$, respectively. For brevity, $f_i$ and $u_i$ will refer to these quantities in the $i$th trajectory.
	
	The ROM we seek to learn involves a $d-$dimensional latent state vector $z$ (with $d \ll D$), whose dynamics are governed by the following system of SDEs,
	\begin{equation} \label{eqn:romsde}
		\begin{aligned}
			\text{d}z = \psi_{\theta} (z, t; \mu, f(t)) 
			\text{d}t + 
			\Psi_{\theta}(t; \mu) \text{d}\beta,\qquad
			z(0) \sim p_\theta^{\rm enc} (z_0 \;|\;u_0(\mu)),
		\end{aligned}
	\end{equation}
	where $\psi_\theta: \real{d} \times [0, T]  \times  \mathcal{F}_{\mu} \times \mathbb{R}^{N_f} \rightarrow \real{d}$ and $\Psi_\theta: [0, T] \times \mathcal{F}_\mu \rightarrow \real{d\times d}$ denote the drift function and dispersion matrix, respectively, and $\beta$ denotes a $d-$dimensional Wiener process. The initial condition of the latent state is specified by the probabilistic encoder $p_\theta^{\rm enc} (z_0 \;|\;u_0(\mu))$, where $z_0 \in \mathbb{R}^d$, and $u_0(\mu) \in \mathbb{R}^D$ denotes the initial condition of the full-order QoI. In addition, we introduce a probabilistic decoder $p_\theta^{\rm dec} (u(t; \mu, f) \;|\; z(t; \mu, f))$ that maps the latent state to the full-order QoI, where  $z(t;\mu,f) \in \mathbb{R}^d$ refers to the solution of \eqref{eqn:romsde} at time $t$ for a given $\mu \in \mathcal{F}_\mu$ and $f \in L^2( [0,T]; \mathbb{R}^{N_f})$.\footnote{In practical settings, where the forcing is only specifed/available at discrete time-stamps, $f$ can be defined as a continuous-time interpolant of these samples.} We use the symbol $\theta$ to denote the learnable parameters of the  encoder, decoder, drift and dispersion models.
	
	A graphical overview of our stochastic ROM framework is presented in Figure~\ref{fig:method}. The framework consists of three parametrized components: a latent SDE model, a probabilistic encoder, and a probabilistic decoder. The drift and diffusion functions of the latent SDE can be parametrized using feedforward neural networks, yielding a ROM in the format of neural SDEs (NSDEs), while the probabilistic encoder and decoder can be parametrized using convolutional neural networks (CNNs). Our framework extends beyond these choices and can accommodate other neural network architectures, including graph neural networks, recurrent neural networks, and transformers. Moreover, our framework also allows for the incorporation of physics-informed parametrizations of the latent SDE; the drift function can be parametrized as $\psi_\theta = \psi^{\rm p} + \psi^{\rm c}_\theta$, where $\psi^{\rm p}$ is a known physics-based prior (for example, obtained by Galerkin projection~\cite{LeeCarlberg2020}) and $\psi^{\rm c}_\theta$ is a parametrized correction term. If the goal is to learn interpretable latent dynamics, the drift function can be parametrized using a basis function expansion combined with a sparsity-inducing prior~\cite{Course_2023_State_Estimation}.
	
	To learn the posterior distributions of the latent state and model parameters, we develop an efficient amortized SVI scheme. We first specify the prior distributions and variational approximations for the latent state and model parameters. We then formulate an evidence lower bound (ELBO) that, when maximized, minimizes the Kullback-Leibler (KL) divergence between the approximate and true posteriors. We show how the ELBO can be reparametrized to eliminate the need for a forward solver during training. Finally, we develop an amortization strategy that decouples the computational cost and memory requirements of evaluating the ELBO from the dataset size. After training, the stochastic ROM can be used to make probabilistic predictions for new parameter settings and forcing conditions as illustrated in Figure~\ref{fig:method}.
	
	\subsection{Priors and variational approximations} \label{sec:prior}
	
	In our ROM framework, the latent state $z$ is a continuous-time stochastic process whose prior is defined by the solution of the nonlinear SDE in \eqref{eqn:romsde}. Our approach defines the variational approximation of the latent state, $q_{\phi}(z \ |\ t; \mu, f)$, implicitly through the solution of the linear SDE
	\begin{equation} \label{eqn:mgp}
		\begin{aligned}
			\text{d}z = \left(-A_{\phi}(t; \mu, f(t)) z + b_{\phi}(t; \mu, f(t))\right)\text{d}t + \Psi_{\theta}(t; \mu) \text{d}\beta,\quad z(0) \sim \mathcal{N}(m_0(\mu), S_0(\mu)),
		\end{aligned}
	\end{equation}
	where $A_{\phi}(t; \mu, f(t)) \in \mathbb{R}^{d \times d}$, $b_{\phi} (t;\mu, f(t)) \in \mathbb{R}^d$, $m_0(\mu) \in \mathbb{R}^d$, and $S_0(\mu) \in \mathbb{R}^{d\times d}$. While this linear SDE representation of $q_\phi$ might seem limiting, it serves purely as a variational approximation during training -- the actual ROM predictions employ the nonlinear SDE in~\eqref{eqn:romsde}.
	
	For the model parameters $\theta$, we may additionally specify a prior $p(\theta)$ and corresponding variational approximation $q_\phi(\theta)$. For instance, when seeking interpretable models, we can parametrize the drift function as a linear combination of basis functions and specify a sparsity-inducing prior on its coefficients~\cite{Course_2023_State_Estimation}. Similarly, we can specify a log-normal prior to constrain the variance of the decoder $p^{\rm dec}_{\theta}(u\ |\ z)$. The variational parameters $\phi$ encompass both the latent state and model parameter posterior approximations. We note that point estimates can also be learned for some or all components of $\theta$ in place of variational approximations; see Section~\ref{sec:priorfree} for details.
	
	\subsection{Evidence lower bound (ELBO)} \label{sec:elbo}
	
	The ELBO is a tractable lower bound on the evidence that serves as an objective function for estimating the variational parameters. Following the derivation in~\cite{Archambeau_2007_Variational_Inference_Diffusion, Course_2023_State_Estimation}, the ELBO for the ROM learning problem can be written as follows:
	\begin{align}\label{eqn:orig_elbo}
		\mathcal{L}(\phi) = & \underbrace{\sum_{i=1}^{N_T} \sum_{j=1}^{N_i} 
			\underset{ z_{i,j}, \theta }{\mathbb{E}}
			\left [ \log p_{\theta}^{\rm dec} \left ( u_{i}(t_{i,j};  \mu_i, f_i) | z_{i,j} \right )  \right ]}_{\text{Log-likelihood term}}\nonumber \\
		- & \underbrace{\sum_{i=1}^{N_T}\frac{1}{2} \int_0^{T} 
			\underset{z_i(t), \theta}{\mathbb{E}}
			\left | \left | r_{\theta, \phi} (z_i(t), t, \mu_i, f_i(t)) \right | \right |_{C_{\theta}(t;\mu_i)}^2 \text{d}t}_{\text{Drift residual term}}
		- \underbrace{D_{\textrm{KL}}\left(q_\phi\left(\theta\right) \|\ p\left(\theta\right)\right)}_{\text{Prior KL term}}, \\
		\textrm{where}\ &\ r_{\theta, \phi}(z_i(t), t, \mu_i, f_i(t)) = -A_\phi (t; \mu_i, f_i(t)) z_i(t) + b_\phi (t; \mu_i, f_i(t)) - \psi_\theta (z_i(t), t; \mu_i, f_i(t)),\nonumber
	\end{align}
	with $z_{i,j} \sim q_\phi(z | t_{i,j}; \mu_i, f_{i})$, $z_i(t) \sim q_\phi(z | t; \mu_i, f_i)$, and $\theta \sim q_\phi(\theta)$, where $r_{\theta, \phi}(z_i(t), t, \mu_i, f_i(t))$ denotes the drift residual corresponding to the $i$th trajectory in the training dataset, $D_{\textrm{KL}}\left(q_\phi\left(\theta\right) \|\ p\left(\theta\right)\right)$ is the KL divergence between the variational posterior and the prior on $\theta$, and $C_{\theta}(t;\mu_i) = (\Psi_\theta(t;\mu_i) \Psi^T_\theta(t;\mu_i))^{-1}$.

	Evaluating the ELBO requires solving the variational SDE in~\eqref{eqn:mgp} for each training trajectory to compute the expectations over $z_{i,j}$
	in the log-likelihood term and over $z_i$ in the drift residual term. This creates a computational bottleneck, particularly for large training datasets and high-dimensional latent states. The following reparametrization addresses this challenge.
	
	\subsection{Reparametrized ELBO} \label{sec:reparam}
	
	We leverage a reparametrization trick for SDE learning proposed by Course and Nair~\cite{Course_2023_State_Estimation} that eliminates the need for a forward solver when computing the ELBO. We begin by noting that the solution of \eqref{eqn:mgp} is a Markov Gaussian process, i.e., $q_\phi(z\ |\ t; \mu, f ) = \mathcal{N}(m_\phi(t; \mu, f), S_\phi(t; \mu, f))$. The dynamics of the mean $m_\phi(t;\mu,f) \in \mathbb{R}^d$ and covariance $S_\phi(t;\mu,f) \in \mathbb{R}^{d \times d}$ are governed by the ordinary differential equations: $\dot{m}_\phi = -A_\phi m_\phi + b_\phi$~\text{and}~$\dot{S}_\phi = -A_\phi S_\phi - S_\phi A_\phi^T + \Psi_\theta \Psi_\theta^T$, with the initial conditions $m_\phi(0; \mu, f) = m_0(\mu)$ and $S_\phi(0; \mu, f) = S_0(\mu)$~\cite{sarkka2019book}. The key insight from~\cite{Course_2023_State_Estimation} is that by expressing $A_\phi$ and $b_\phi$ in terms of $m_\phi$ and $S_\phi$, the ELBO can be reparametrized in terms of $m_\phi$ and $S_\phi$. 
	
	The reparametrized ELBO retains the structure of \eqref{eqn:orig_elbo}, but introduces two crucial changes. First, the expectations in the ELBO are taken using samples drawn from $\mathcal{N}(m_\phi, S_\phi)$ which does not require a forward solver, i.e., $z_{i,j} \sim \mathcal{N} ( m_\phi(t_{i,j}; \mu_i, f_i), S_\phi (t_{i,j}; \mu_i, f_i) )$, and $z_i(t) \sim \mathcal{N} ( m_\phi(t; \mu_i, f_i), S_\phi (t; \mu_i, f_i) )$. Second, the original drift residual is replaced with the following reparametrized drift residual that depends only on $m_\phi$ and $S_\phi$:
	\begin{equation} \label{eqn:reparam_driftresidual}
		r_{\theta, \phi}(z_i(t), t, \mu_i, f_i) = B \, (m_\phi(t; \mu_i, f_i) - z_i(t)) + \dot{m}_\phi(t; \mu_i, f_i) - 
		\psi_{\theta}(z_i(t),t;\mu_i,f_i),
	\end{equation}
	where $B = {\rm vec}^{-1}\left((S_\phi \oplus S_\phi)^{-1} {\rm vec}(\Psi_\theta \Psi^T_\theta - \dot{S}_\phi )\right)$, $\oplus$ denotes the Kronecker sum, $\mathrm{vec}: \mathbb{R}^{d\times d} \rightarrow \mathbb{R}^{d^2}$ is an operator that maps a matrix to a vector by stacking columns, and $\mathrm{vec}^{-1}\ :\ \mathbb{R}^{d^2} \rightarrow \mathbb{R}^{d \times d}$ denotes its inverse. The time-derivatives of the variational mean and covariance in the reparametrized drift residual are given by
	\begin{equation} \label{eqn:ms_chain}
		\begin{aligned}
			\dot{m}_\phi(t; \mu, f) & = \pp{m_\phi}{t}(t; \mu, f) + \pp{m_\phi}{f}(t; \mu, f) \, \dot{f}(t),\\
			\dot{S}_\phi(t; \mu, f) & = \pp{S_\phi}{t}(t; \mu, f) + \pp{S_\phi}{f}(t; \mu, f) \, \dot{f}(t).
		\end{aligned}
	\end{equation}

	Despite the significant computational advantages of the reparametrized ELBO, two limitations remain. 
	First, similar to~\eqref{eqn:orig_elbo}, the reparametrized ELBO does not depend on the encoder $p_\theta^{\rm enc} (z_0 \;|\;u_0(\mu))$, which is necessary for initializing the latent state when making ROM predictions. Second, the variational approximation for the latent state is defined over the entire time interval $[0,T]$ and space of forcings $\real{N_f}$, potentially limiting its effectiveness for modeling complex dynamics over long time horizons. We address these limitations through an amortization strategy.

	\subsection{Amortized SVI with continuous-time encoding} \label{sec:amortized}
	
	We adopt an amortization strategy~\cite{Course_2023_Amortized_Reparameterization}, where the variational distribution of the latent state is learned over partitions of $[0,T]$. Consider the $i$th trajectory where the time-stamps are given by $\{ t_{i,1}, t_{i,2}, \ldots, t_{i,N_i} \}$, where $t_{i,1} = 0$ and $t_{i,N_i} = T$. With window size $M$, we partition the time interval $[0,T]$ into $N_i/M$ non-overlapping subintervals (if $N_i$ is not divisible by $M$, we use a smaller partition size for the last subinterval) as follows:
	\begin{equation} \label{eqn:time_partition}
		\begin{aligned}
			\Big\{ & \left \{
			t_{i,1}^{(1)}, ~ \ldots ~, t_{i,M}^{(1)}\right \}, \left \{ t_{i,1}^{(2)}, ~ \ldots ~, t_{i,M}^{(2)} \right \}, 
			\ldots, \left \{
			t_{i,1}^{(N_i/M)}, ~\ldots~, t_{i,M}^{(N_i/M)} \right \} \Big\},
		\end{aligned}
	\end{equation}
	where we set $t_{i,M}^{(k)} = t_{i,1}^{(k+1)}$ for $k=1,2,\ldots,N_i/M-1$.
	
	The variational distribution of the latent state can be subsequently amortized over the time partitions. More specifically, the variational distribution of the latent state over the $k$th time partition for the $i$th trajectory is written as $q_\phi(z | t; \mu_i, f_i, \mathcal{D}_{i,k}) = \mathcal{N}(m_\phi(t; \mu_i, f_i, \mathcal{D}_{i,k}), S_\phi(t; \mu_i, f_i, \mathcal{D}_{i,k}))$, where $t \in [t_{i,1}^{(k)}, t_{i,M}^{(k)}]$ and $\mathcal{D}_{i,k} := \{ u_i(t_{i,j}^{(k)}; \mu_i, f_i) \}_{j=1}^M$ denotes the FOM QoI observations within the $k$th time partition for the $i$th trajectory. The  amortized ELBO takes the form 
	\begin{equation} \label{eqn:amortized_elbo}
		\begin{aligned}
			\mathcal{L}(\phi) = & \sum_{i=1}^{N_T} 
			\sum_{k=1}^{N_i/M}
			\Bigg( \sum_{j=1}^{M} 
			\underset{ z_{i,j}^{(k)}, \theta }{\mathbb{E}}
			\left [ \log p_{\theta}^{\rm dec} \left ( u_{i}(t_{i,j}^{(k)};  \mu_i, f_i) | z_{i,j}^{(k)} \right )  \right ] \\
			- & \frac{1}{2} \int_{t_{i,1}^{(k)}}^{t_{i,M}^{(k)}}
			\underset{z_i^{(k)} (t), \theta}{\mathbb{E}}
			\left | \left | r_{\theta, \phi} (z_i^{(k)}(t), t, \mu_i, f_i(t)) \right | \right |_{C_{\theta}(t;\mu_i)}^2 \text{d}t \Bigg) - D_{\textrm{KL}}\left(q_\phi\left(\theta\right) \|\ p\left(\theta\right)\right),
		\end{aligned}
	\end{equation}
	with $z_{i,j}^{(k)} \sim \mathcal{N} ( m_\phi(t_{i,j}^{(k)}; \mu_i, f_i, \mathcal{D}_{i,k}), S_\phi (t_{i,j}^{(k)}; \mu_i, f_i, \mathcal{D}_{i,k}) )$, $z_i^{(k)}(t) \sim \mathcal{N} ( m_\phi(t; \mu_i, f_i, \mathcal{D}_{i,k}), S_\phi (t; \mu_i, f_i, \mathcal{D}_{i,k}) )$, and $\theta \sim q_\phi(\theta)$.
	
	To construct the latent variational distribution, we apply a two-step process within each partition. First, the QoI samples from the partition are passed as inputs to the encoder $p_\theta^{\rm enc}(z\ |\ u)$. Then, we apply a deep-kernel interpolation scheme to these encoded values, yielding a continuous-time representation of $m_\phi(t)$ and $S_\phi(t)$.
	
	We use a diagonal parametrization of $S_{\phi}$ to ensure parameter efficiency for large values of $d$, and denote the diagonal entries of $S_\phi$ by the $d-$dimensional vector $s_\phi$. We next define a CNN that maps the QoI snapshot to the latent state mean and variance, i.e., $(m_\phi(t;\mu,f), \log(s_\phi(t;\mu,f))) = \mathcal{Y}(u(t;\mu,f))$. A log transformation is used to ensure that all entries of $s_\phi$ are strictly positive. It is worth noting here that, while making predictions, this CNN will serve as the encoder for the latent initial condition of our ROM in \eqref{eqn:romsde} given the QoI initial condition, i.e., $p_\theta^{\rm enc} (z_0 \;|\;u_0(\mu)) = \mathcal{N}(m_0(\mu),S_0(\mu))$, where  $S_0(\mu) = \text{diag}(s_0(\mu))$ and $(m_0(\mu), \log s_0(\mu)) = \mathcal{Y}(u_0(\mu))$. The CNN $\mathcal{Y}$ takes the QoI snapshot $u_i(t_{i,j}^{(k)};\mu_i,f_i)$ as input and its outputs are the mean and log-variance of the Gaussian distribution for $z_{i,j}^{(k)}$. This parametrization enables all the expectations in the log-likelihood term of the ELBO in~\eqref{eqn:amortized_elbo} to be evaluated. 
	
	The continuous-time representation of the variational parameters $m_\phi$ and $S_\phi$ over the $k$th time partition for each trajectory must be differentiable to permit evaluation of the reparametrized drift residual in~\eqref{eqn:reparam_driftresidual}. To proceed further, we first evaluate the CNN $\mathcal{Y}$ over all snapshots of the QoI in the $k$th time partition for the $i$th trajectory. This leads to the matrix of feature vectors $H_m = [ m_{i,1}^{(k)},~m_{i,2}^{(k)},~ \cdots,~ m_{i,M}^{(k)} ]^T \in \mathbb{R}^{M \times d}$ and $H_s = [ \log s_{i,1}^{(k)},~ \log s_{i,2}^{(k)},~\cdots, ~\log s_{i,M}^{(k)}	]^T \in \mathbb{R}^{M \times d}$, where $(m_{i,j}^{(k)}, \log s_{i,j}^{(k)} ) = \mathcal{Y}(u_i(t_{i,j}^{(k)}; \mu_i, f_i))$, $j=1,2,\ldots,M$. We then apply the deep kernel interpolation scheme from~\cite{Course_2023_Amortized_Reparameterization} to the feature vectors in $H_m$ and $H_s$ to arrive the following continuous-time representations of $m_\phi$ and $S_\phi$:
	\begin{equation} \label{eqn:ms_kernel}
		\begin{aligned}
			m_\phi(t; \mu_i, f_i) & = K_{M}(t) 
			\left( K_{MM} + \sigma^2  I\right)^{-1} H_m, \\
			\log s_\phi(t; \mu_i, f_i) & = K_{M}(t) 
			\left( K_{MM} + \sigma^2 I\right)^{-1} H_s,
		\end{aligned}
	\end{equation}
	where $t \in [t_{i,1}^{(k)}, t_{i,M}^{(k)}]$, $K_M(t)$ is a row vector of length $M$ whose entries are given by $\kappa(t, t_{i,j}^{(k)}), j=1,2,\ldots,M$, $K_{MM}$ is an $M\times M$ matrix whose entries are $\kappa(t_{i,j}^{(k)}, t_{i,j'}^{(k)}), j=1,2,\ldots,M,j'=1,2,\ldots,M$, $I$ is the identity matrix of size $M$, and $\kappa: \mathbb{R} \times \mathbb{R} \rightarrow \mathbb{R}$ denotes a deep kernel that is defined as $\kappa(t_1, t_2) = \sigma_f \exp\left(-\|\varphi(t_1) - \varphi(t_2)\|^2/(2\ell^2)\right)$, where $\varphi: \real{} \rightarrow \real{}$ is a deep feedforward neural network. The parameter $\sigma_f$ controls the amplitude of the kernel and $\ell$ determines the length scale. The parameters $\sigma, \sigma_f, \ell \in \real{+}$ are estimated along with other model parameters during training.
	
	Note that following~\eqref{eqn:ms_kernel}, the effect of forcing on $m_\phi$ and $S_\phi$ is contained in the $H_m$ and $H_s$ terms respectively, which are constant. The only time-dependency is via the kernel vector $K_M(t)$, which makes the chain rule expansion in~\eqref{eqn:ms_chain} unnecessary.
	
	\subsection{Gradient estimation} \label{sec:gradient}
	
	To estimate the gradient of the ELBO with respect to variational parameters $\phi$, we use the standard reparametrization trick~\cite{Kingma_2022_Autoencoding_Variational_Bayes}. We express the reparametrized latent state as $z(t; \mu,f) = \mathcal{T}(t,\gamma,\phi)= m_\phi(t; \mu, f) + (S_\phi(t; \mu, f))^{1/2}\gamma$, where $\gamma \sim \mathcal{N}(0, I)$. Similarly, we express the reparametrized model parameters as $\theta = \mathcal{Q}(\xi, \phi)$, where $\xi \sim \mathcal{N}(0, I)$. The drift residual integral is approximated using a Monte Carlo scheme with time samples drawn from $\mathcal{U}[t_{i,1}^{(k)}, t_{i,M}^{(k)}]$. Taking $R$ samples of $\gamma$, $L$ time samples, and one sample of $\xi$, we obtain the following unbiased estimate of the ELBO gradient over the $k$th partition of the $i$th trajectory.
	\begin{align} \label{eqn:elbo_grad}
		& \nabla_{\phi} \mathcal{L}(\phi) \approx \frac{1}{R}\sum_{r=1}^{R}
		\Bigg( \sum_{j=1}^{M}
		\nabla_{\phi}\log p_{\theta}^{\rm dec} \left ( u_{i}(t_{i,j}^{(k)};  \mu_i, f_i)\ |\ z_{i,j,r}^{(k)}\right ) \nonumber \\
		& - \frac{t_{i,M}^{(k)}-t_{i,1}^{(k)}}{2L} 
		\sum_{\ell=1}^{L}
		\nabla_{\phi}\left\| r_{\theta, \phi} \left(z_{\ell,r}, \tau_\ell, \mu_i, f_i(\tau_{\ell})\right) \right\|_{C_{\theta}(\tau_{\ell};\mu_i)}^2 \Bigg) - \nabla_{\phi} D_{\textrm{KL}}\left(q_\phi\left(\theta \right) \|\ p\left(\theta \right)\right),
	\end{align}
	where $z_{i,j,r}^{(k)} = \mathcal{T}(t_{i,j}^{(k)},\gamma_r,\phi)$, $z_{\ell,r} = \mathcal{T}(\tau_\ell,\gamma_{r},\phi)$, the indices $i$ and $k$ are sampled uniformly from $\{1,2, \dots, N_T\}$ and $\{1,2, \dots, N_i/M\}$ respectively, $\theta = \mathcal{Q}(\xi, \phi)$ where $\xi$ is a single sample drawn from $\mathcal{N}(0, I)$, $\gamma_1, \gamma_2, \dots, \gamma_R \sim \mathcal{N}(0, I)$, and $\tau_1, \tau_2, \dots, \tau_L \sim \mathcal{U}[t_{i,1}^{(k)}, t_{i,M}^{(k)}]$.
		 
	The key advantage of our reparametrization trick and amortized representation is that the computational complexity and memory requirements of our gradient estimator are independent of:  (i)~training dataset size (number of trajectories), (ii)~time series length (number of samples per trajectory), and (iii)~stiffness of either the underlying dynamics or the approximate SDE. 
    More specifically, the computational complexity scales linearly with the sampling parameters $R$, $M$, and $L$. Furthermore, our approach enables massively parallel computation since each gradient term can be evaluated independently. This contrasts sharply with the original ELBO in \eqref{eqn:orig_elbo}, which depends on a strictly sequential SDE forward solver.
	
	\subsection{Point estimates of model parameters} \label{sec:priorfree}
	
	For the case when a point estimate is sought for the parameter vector $\theta$ rather than a distribution $q_\phi(\theta)$, the amortized ELBO in~\eqref{eqn:amortized_elbo} simplifies to
	\begin{eqnarray} \label{eqn:priorfree_amortized_elbo}
		\mathcal{L}(\theta, \phi) & = & \sum_{i=1}^{N_T} 
		\sum_{k=1}^{N_i/M}
		\Bigg( \sum_{j=1}^{M} 
		\underset{ z_{i,j}^{(k)} }{\mathbb{E}}
		\left [ \log p_{\theta}^{\rm dec} \left ( u_{i}(t_{i,j}^{(k)};  \mu_i, f_i)\ |\ z_{i,j}^{(k)} \right )  \right ] \nonumber \\
		& - & \frac{1}{2} \int_{t_{i,1}^{(k)}}^{t_{i,M}^{(k)}}
		\underset{z_i^{(k)} (t)}{\mathbb{E}}
		\left | \left | r_{\theta, \phi} (z_i^{(k)}(t), t, \mu_i, f_i(t)) \right | \right |_{C_{\theta}(t;\mu_i)}^2 \text{d}t \Bigg),
	\end{eqnarray}
	with $z_{i,j}^{(k)} \sim \mathcal{N} ( m_\phi(t_{i,j}^{(k)}; \mu_i, f_i, \mathcal{D}_{i,k}), S_\phi (t_{i,j}^{(k)}; \mu_i, f_i, \mathcal{D}_{i,k}) )$ and $z_i^{(k)}(t) \sim \mathcal{N} ( m_\phi(t; \mu_i, f_i, \mathcal{D}_{i,k}), S_\phi (t; \mu_i, f_i, \mathcal{D}_{i,k}) )$.
	Note that the term $D_{KL}(q_\phi(\theta) || p(\theta))$ is no longer present in the ELBO and we no longer take expectations over $\theta$ in the log-likelihood and drift residual terms.
	The unbiased estimate of the ELBO gradient over the $k$th partition of the $i$th trajectory then takes the following form (analogous to~\eqref{eqn:elbo_grad}):
	\begin{eqnarray} \label{eqn:priorfree_elbo_grad}
		\nabla_{\theta,\phi} \mathcal{L}(\theta,\phi) & \approx & 
		\frac{1}{R}\sum_{r=1}^{R}
		\Bigg( \sum_{j=1}^{M}
		\nabla_{\theta,\phi}\log p_{\theta}^{\rm dec} \left ( u_{i}(t_{i,j}^{(k)};  \mu_i, f_i)\ |\ \mathcal{T}(t_{i,j}^{(k)},\gamma_r,\phi) \right ) \nonumber \\
		& - & \frac{t_{i,M}^{(k)}-t_{i,1}^{(k)}}{2L} 
		\sum_{\ell=1}^{L}
		\nabla_{\theta,\phi}\left\| r_{\theta, \phi} \left(\mathcal{T}(\tau_\ell,\gamma_{r},\phi), \tau_\ell, \mu_i, f_i(\tau_{\ell})\right) \right\|_{C_{\theta}(\tau_{\ell};\mu_i)}^2 \Bigg),
	\end{eqnarray}
	where the indices $i$ and $k$ are sampled uniformly from $\{1,2, \dots, N_T\}$ and $\{1, 2,\dots, N_i/M\}$, respectively, $\gamma_1, \gamma_2, \dots, \gamma_R \sim \mathcal{N}(0, I)$, and $\tau_1, \tau_2, \dots, \tau_L \sim \mathcal{U}[t_{i,1}^{(k)}, t_{i,M}^{(k)}]$. We note that since $\theta$ is no longer a random variable, the reparameterization in terms of $\mathcal{Q}(\xi, \phi)$ is no longer necessary. Similar to~\eqref{eqn:elbo_grad}, we express the reparametrized latent state as $z(t; \mu,f) = \mathcal{T}(t,\gamma,\phi)= m_\phi(t; \mu, f) + (S_\phi(t; \mu, f))^{1/2}\gamma$, where $\gamma \sim \mathcal{N}(0, I)$.

	We now consider the case when point estimates are only sought for a subset of the model parameters in $\theta$. Let $\widehat{\theta}$ denote the subset of parameters for which point estimates are sought 
	while $\widetilde{\theta}$ denotes the set of model parameters for 
	which a prior and variational approximation is specified (i.e., $\theta$ is decomposed into $(\widetilde{\theta},\widehat{\theta})$). In this case, training the model yields a posterior distribution over $\widetilde{\theta}$, parametrized by $\phi$, and a point estimate of $\widehat{\theta}$. In this setting, the ELBO takes the form
	\begin{align} \label{eqn:partialprior_amortized_elbo}
		\mathcal{L}(\widehat{\theta}, \phi) & = \sum_{i=1}^{N_T} 
		\sum_{k=1}^{N_i/M}
		\Bigg( \sum_{j=1}^{M} 
		\underset{ z_{i,j}^{(k)}, \widetilde{\theta} }{\mathbb{E}}
		\left [ \log p_{\theta}^{\rm dec} \left ( u_{i}(t_{i,j}^{(k)};  \mu_i, f_i)\ |\ z_{i,j}^{(k)} \right )  \right ] \nonumber \\
		& - \frac{1}{2} \int_{t_{i,1}^{(k)}}^{t_{i,M}^{(k)}}
		\underset{z_i^{(k)} (t), \widetilde{\theta}}{\mathbb{E}}
		\left | \left | r_{\theta, \phi} (z_i^{(k)}(t), t, \mu_i, f_i(t)) \right | \right |_{C_{\theta}(t;\mu_i)}^2 \text{d}t \Bigg) - D_{\textrm{KL}}\left(q_\phi(\widetilde{\theta}) \|\ p(\widetilde{\theta})\right),
	\end{align}
	with $z_{i,j}^{(k)} \sim \mathcal{N} ( m_\phi(t_{i,j}^{(k)}; \mu_i, f_i, \mathcal{D}_{i,k}), S_\phi (t_{i,j}^{(k)}; \mu_i, f_i, \mathcal{D}_{i,k}) )$ and $z_i^{(k)}(t) \sim \mathcal{N} ( m_\phi(t; \mu_i, f_i, \mathcal{D}_{i,k}), S_\phi (t; \mu_i, f_i, \mathcal{D}_{i,k}) )$.
	Note that the expectations in the log-likelihood and drift residual terms are now over $\widetilde{\theta}$, and the KL term is  defined between the prior and variational approximation of $\widetilde{\theta}$.
	
	To obtain an unbiased estimator for the gradients of the ELBO with respect to the variational parameters $\phi$, and model parameters $\widehat{\theta}$, we reparametrize $\widetilde{\theta}$ as $\widetilde{\theta}=Q(\xi, \phi)$ with $\xi \sim \mathcal{N}(0, I)$ and express the reparametrized latent state as $z(t; \mu,f) = \mathcal{T}(t,\gamma,\phi)= m_\phi(t; \mu, f) + (S_\phi(t; \mu, f))^{1/2}\gamma$ with $\gamma \sim \mathcal{N}(0, I)$.  This  yields
	\begin{eqnarray} \label{eqn:partialprior_elbo_grad}
		\nabla_{\widehat{\theta},\phi} \mathcal{L}(\widehat{\theta},\phi) & \approx & 
		\frac{1}{R}\sum_{r=1}^{R}
		\Bigg( \sum_{j=1}^{M}
		\nabla_{\widehat{\theta},\phi}\log p_{\theta}^{\rm dec} \left ( u_{i}(t_{i,j}^{(k)};  \mu_i, f_i)\ |\ \mathcal{T}(t_{i,j}^{(k)},\gamma_r,\phi) \right ) \nonumber \\
		& - & \frac{t_{i,M}^{(k)}-t_{i,1}^{(k)}}{2L} 
		\sum_{\ell=1}^{L}
		\nabla_{\widehat{\theta},\phi}\left\| r_{\theta, \phi} \left(\mathcal{T}(\tau_\ell,\gamma_{r},\phi), \tau_\ell, \mu_i, f_i(\tau_{\ell})\right) \right\|_{C_{\theta}(\tau_{\ell};\mu_i)}^2 \Bigg) \nonumber \\
		& - & \nabla_{\phi} D_{\textrm{KL}}\left(q_\phi\left(\widetilde{\theta} \right) \|\ p\left(\widetilde{\theta} \right)\right),
	\end{eqnarray}
	where the indices $i$ and $k$ are sampled uniformly from $\{1,2, \ldots, N_T\}$ and $\{1,2, \ldots, N_i/M\}$, respectively, $\widetilde{\theta} = \mathcal{Q}(\xi, \phi)$ where $\xi$ is a single sample drawn from $\mathcal{N}(0, I)$, $\gamma_1, \gamma_2, \ldots, \gamma_R \sim \mathcal{N}(0, I)$, and $\tau_1, \tau_2, \ldots, \tau_L \sim \mathcal{U}[t_{i,1}^{(k)}, t_{i,M}^{(k)}]$.
	
	\subsection{Probabilistic ROM predictions}
	
	To generate predictions with the trained ROM, we require an initial condition for the full-order QoI $u_0$, system parameters $\mu$, forcing function $f$, and a simulation time $T$. The prediction process involves the following steps:
	\begin{enumerate}
		\item Sample latent initial condition from encoder, $z(0) \sim p_{\theta}^{\rm enc}(z_0\ |\ u_0)$,
		\item Evolve latent state by integrating \eqref{eqn:romsde}, $z(T) \sim \int_0^T {\psi}_{{\theta}}(z, t; \mu, f(t)) \text{d}t + \int_0^T {\Psi}_{{\theta}}(t; \mu) \text{d}{\beta}$,
		\item Generate QoI prediction by sampling from decoder, $u(T) \sim p_{\theta}^{\rm dec}(u(T)\ |\ z(T))$.
	\end{enumerate}
	This procedure forms the basis for our model evaluation in the numerical studies that follow.
	
	\section{Results} \label{sec:examples}
	
	To comprehensively evaluate the efficacy and versatility of our proposed framework for learning stochastic ROMs, we conduct numerical studies on three challenging test problems.
	These range from learning sparse, interpretable dynamics in unforced systems, to handling complex parametrized and forced dynamics, and finally, to assessing scalability on a high-dimensional controlled fluid flow problem. The first test case is a non-parametrized, unforced reaction-diffusion system~\cite{Champion_2019_Data_Driven_Coordinates} that enables an evaluation of our approach's ability to infer sparse latent dynamics without requiring time-derivative observations. The second test case is a forced, parametrized one-dimensional Burgers' equation problem that introduces the challenge of learning ROMs that generalize across parameter spaces and forcing conditions, a scenario commonly encountered in many engineering applications. Finally, we study a controlled fluid flow problem~\cite{Rabault_2019_Flow_Control_DRL} to demonstrate the scalability and practical applicability of our method to large-scale systems.
	In each study, ROMs are trained using a set of sample trajectories and their performance is rigorously assessed on a separate, unseen test set using the error metric: $\varepsilon(t) = \|{u}(t) - \overline{{u}}(t)\|/\|{u}(t)\|$, where $\overline{{u}}(t)$ is the mean ROM prediction and ${u}(t)$ denotes the true QoI. 
	
	We benchmark our approach against four established data-driven ROM techniques, spanning both neural differential equation models and sparse, interpretable model discovery frameworks. We consider two variants of latent SINDy, a framework for inferring sparse interpretable models~\cite{Conti_2023_ROM_Parametrized_SINDy,Nicolaou_2023_Data_Driven_Parametrized}. We implement these using the PySINDy library~\cite{De_Silva_2020_PySINDy,Kaptanoglu_2022_PySINDy}: (i)~POD-SINDy, employing POD for dimensionality reduction, and (ii)~AE-SINDy, utilizing a non-linear autoencoder. In addition, we also benchmark against parametrized neural ordinary differential equations (PNODEs) following~\cite{Lee_2021_Parameterized_Neural_ODE_ROM} (implemented with the \verb|torchdiffeq| library~\cite{Chen_2018_NODEs}), and an extension thereof into parametrized neural stochastic differential equations (PNSDEs) (implemented with the \verb|torchsde| library~\cite{Li_2020_Scalable_Gradients_VI_SDE}). Following Chen et al~\cite{Chen_2018_NODEs} and Rubanova et al.~\cite{Rubanova_2019_NODE}, we use an evidence lower bound as our loss; we detail the PNODE/PNSDE formulations in Section~\ref{sec:pnodepnsde}. Note that the training of PNODEs and PNSDEs relies on a forward solver, a bottleneck our approach aims to eliminate. Furthermore, PNODE/PNSDE architectures do not directly accommodate time-varying forcing functions; thus, for forced problems, we adopt the common practice of parametrizing the forcing via its characteristic parameters or POD coefficients.
	The PNODE and PNSDE baselines are trained using full-trajectory batches, a common practice for these methods. All models were trained for a comparable number of epochs to facilitate a fair comparison; for additional details see Section~\ref{sec:testdetails}.
	
	\begin{table}[t]
		\centering
		\caption{Reaction-diffusion: model comparison over mean error $\varepsilon_\mu$, standard deviation of error $\varepsilon_\sigma$, and training time $T_{\rm tr}$ in hours. PA: present approach. AE-SINDy$^1$ is trained in a two-step manner using PySINDy. AE-SINDy$^2$ is trained in an end-to-end manner using the architecture and implementation from Champion et al.}
		\label{tab:reactiondiffusion_comparison}
		\begin{tabular}{c|cccc|cccc}
			\toprule
			& \multicolumn{4}{c|}{Noiseless dataset}
			& \multicolumn{4}{c}{Noisy dataset} \\
			\midrule
			& & POD- & AE- & AE- & & POD- & AE- & AE- \\
			Metric & PA & SINDy & SINDy$^1$ & SINDy$^2$ & PA & SINDy & SINDy$^1$ & SINDy$^2$ \\
			\midrule
			$\varepsilon_\mu$ & $0.036$ & $0.140$ & $1.141$ & $0.126$ & $0.052$ & $0.142$ & $1.212$ & $0.130$ \\
			$\varepsilon_\sigma$ & $0.015$ & $0.003$ & $0.652$ & $0.006$ & $0.018$ & $0.003$ & $0.620$ & $0.008$ \\
			$T_{\rm tr}$ & $3.698$ & $0.009$ & $0.629$ & $1.699$ & $3.630$ & $0.014$ & $0.613$ & $1.603$ \\
			\bottomrule
		\end{tabular}
	\end{table}
	
	\begin{figure}[t]
		\centering
		\parbox{0.5\textwidth}{\centering \large Noiseless Data}%
		\parbox{0.5\textwidth}{\centering \large Noisy Data}%
		
		\includegraphics[width=0.5\linewidth]{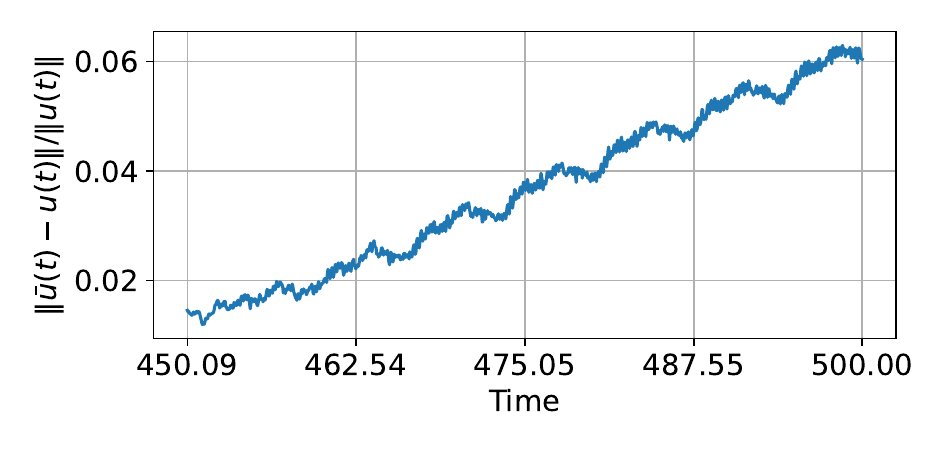}%
		\includegraphics[width=0.5\linewidth]{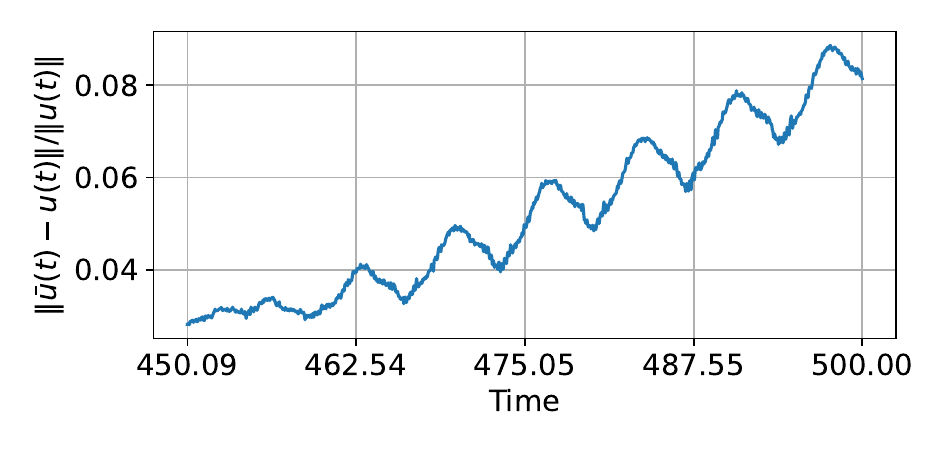}%
		
		\includegraphics[width=0.5\textwidth]{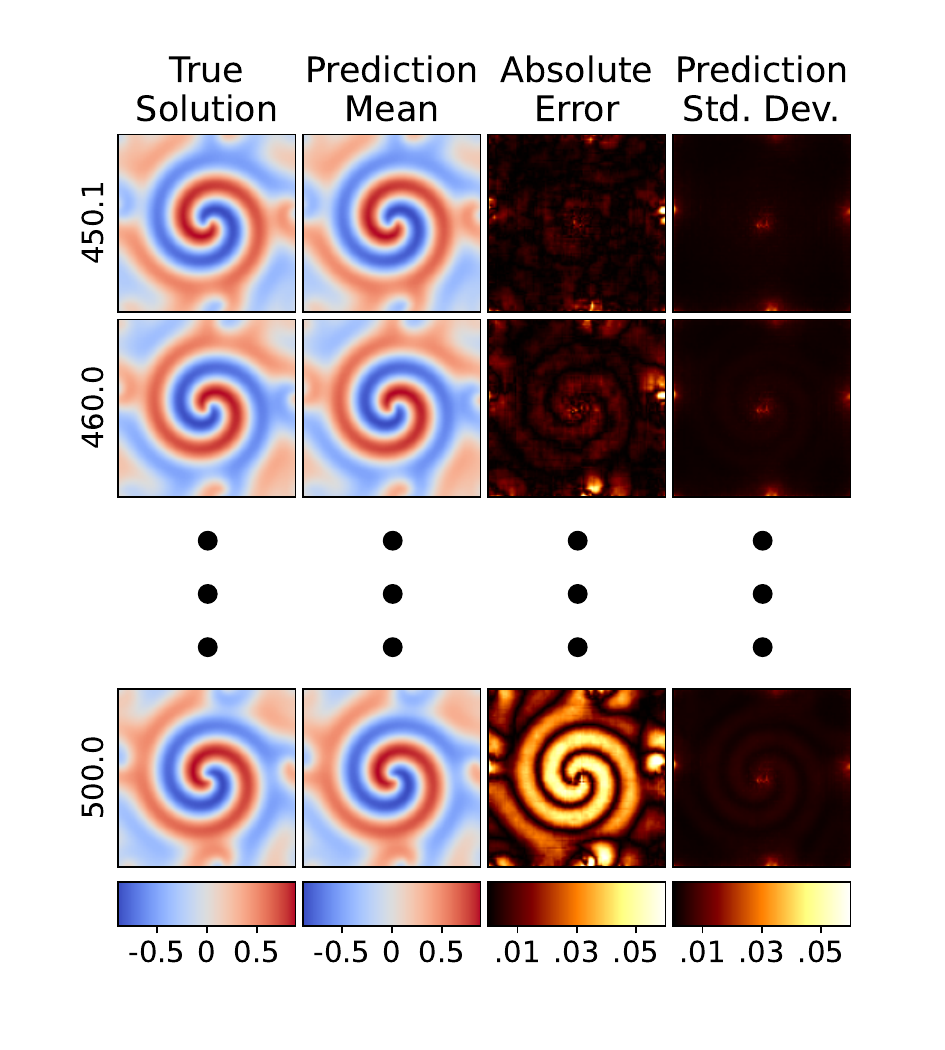}%
		\includegraphics[width=0.5\textwidth]{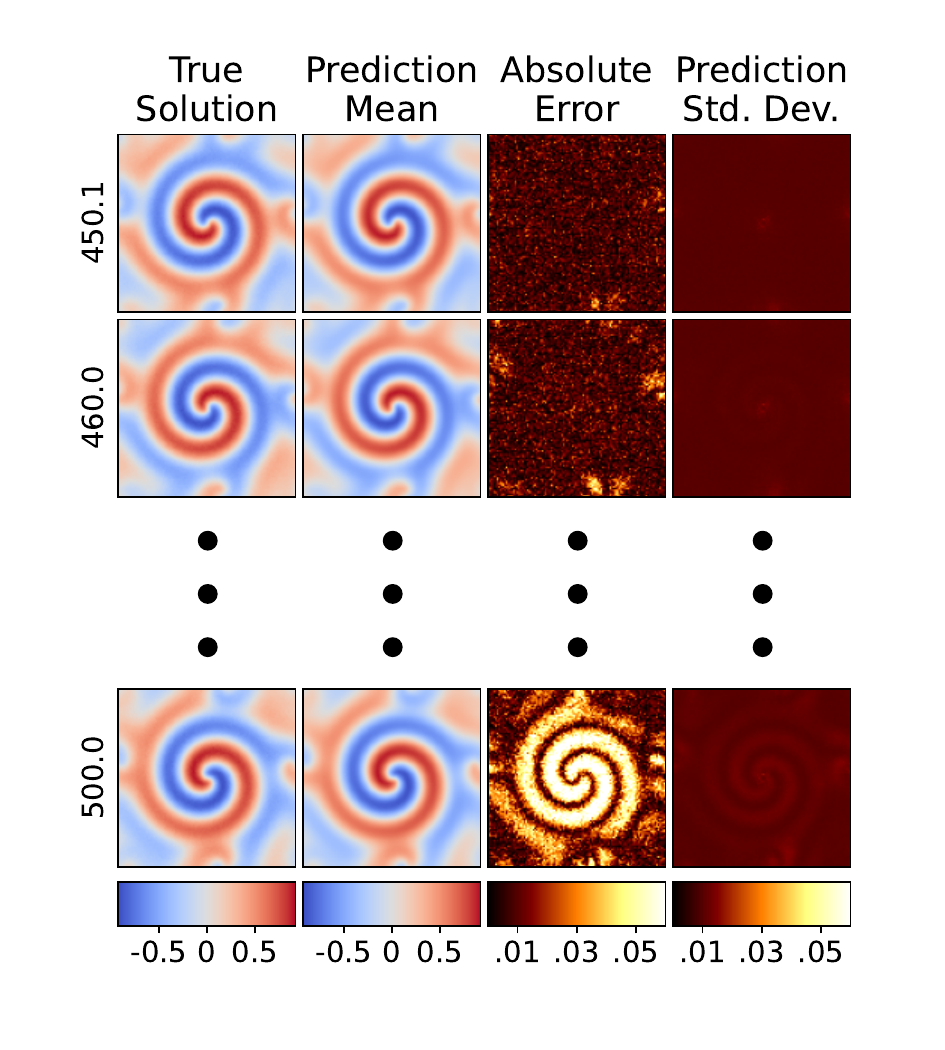}%
		
		\caption{Reaction-diffusion: error evolution and predictions over time. The top row illustrates prediction error on the test interval for noiseless (left) and noisy (right) data. The bottom half shows the true solution, prediction mean, absolute error, and prediction standard deviation. See movies S1-S2.}\label{fig:reactiondiffusion}
	\end{figure}
	
	\subsection*{Example 1: Reaction-diffusion system}
	
	We first consider a reaction-diffusion system previously studied by Champion et al.~\cite{Champion_2019_Data_Driven_Coordinates}. Details of the problem setup are given in Section~\ref{sec:reacdiffdetails}. The spatial domain is discretized using a grid with $D=100 \times 100$ equispaced points, and the time domain is discretized using $9999$ steps. We consider a single trajectory with fixed parameters and no forcing. The time interval is partitioned as follows: the interval $[0, 400]$ constitutes the training set, the next interval $(400, 450]$ is the validation set, and the final $(450, 500]$ is the test set. We set up the training data in a manner identical to Champion et al., except that whereas they corrupt the data (both $v$ and its time derivative $v_t$, independently) using zero-mean Gaussian noise with standard deviation $10^{-6}$, we consider two cases: a noiseless case, and a noisy case with standard deviation $10^{-2}$.
	
	Following Champion et al., we consider a two-dimensional latent state and parametrize the drift function $\psi_\theta$ as a third-order polynomial. The results obtained using different methods are summarized in Table~\ref{tab:reactiondiffusion_comparison}. The table highlights the importance of the training strategy for latent SINDy models. Specifically, AE-SINDy$^1$, trained in a two-step manner, performs poorly; its autoencoder, optimized independently of the SINDy dynamics model, likely yields latent trajectories that are difficult for SINDy to represent accurately. In contrast, POD-SINDy, despite using a simpler linear encoder, achieves better results than AE-SINDy$^1$, possibly due to the inherent smoothness and regularity often associated with POD modes. The end-to-end training of AE-SINDy$^2$ demonstrates a significant improvement over both, underscoring the benefit of jointly optimizing the encoder and the latent dynamics. 
	
	Our approach (PA) provides better accuracy compared to all SINDy-based methods on both datasets. Notably, this higher accuracy is achieved using only state measurements, whereas all SINDy variants require time-derivative information. The test predictions of our stochastic ROM are shown in Figure~\ref{fig:reactiondiffusion} for both noiseless and noisy data. It can be seen that the stochastic ROM is able to capture the dynamics well, even in the presence of noise. The error growth over time is approximately linear, which can be attributed to the ROM drifting out of phase with the true trajectory.
	
	\begin{table}[t]
		\begin{minipage}{0.58\textwidth}
			\caption{Burgers' equation: model comparison for varying ROM dim. $d$ over mean error $\varepsilon_\mu$, standard deviation of error $\varepsilon_\sigma$, and training time $T_{\rm tr}$ in hours. PA: present approach.}
			\label{tab:burgers_comparison}
			\begin{tabular}{cc|ccccc}
				\toprule
				& & & POD- & AE- & & \\
				$d$ & Metric & PA & SINDy & SINDy & PNODE & PNSDE \\
				\midrule
				& $\varepsilon_\mu$ & $0.081$ & $0.903$ & $0.961$ & $0.075$ & $0.086$ \\
				$3$ & $\varepsilon_\sigma$ & $0.086$ & $0.084$ & $0.307$ & $0.015$ & $0.26$ \\
				& $T_{\rm tr}$ & $3.352$ & $0.001$ & $0.553$ & $5.982$ & $4.358$ \\
				\midrule
				& $\varepsilon_\mu$ & $0.065$ & $0.719$ & $0.192$ & $0.707$ & $0.718$ \\
				$4$ & $\varepsilon_\sigma$ & $0.028$ & $0.092$ & $0.111$ & $0.004$ & $0.009$ \\
				& $T_{\rm tr}$ & $3.281$ & $0.001$ & $0.552$ & $6.035$ & $4.407$ \\
				\midrule
				& $\varepsilon_\mu$ & $0.067$ & $0.569$ & $0.276$ & $0.162$ & $0.175$ \\
				$5$ & $\varepsilon_\sigma$ & $0.032$ & $0.070$ & $0.425$ & $0.078$ & $0.077$ \\
				& $T_{\rm tr}$ & $3.271$ & $0.004$ & $0.553$ & $4.877$ & $4.559$ \\
				\bottomrule
			\end{tabular}
		\end{minipage}\hfill
		\begin{minipage}{0.4\textwidth}
			\centering
			\vspace{1em}
			\includegraphics[width=\linewidth]{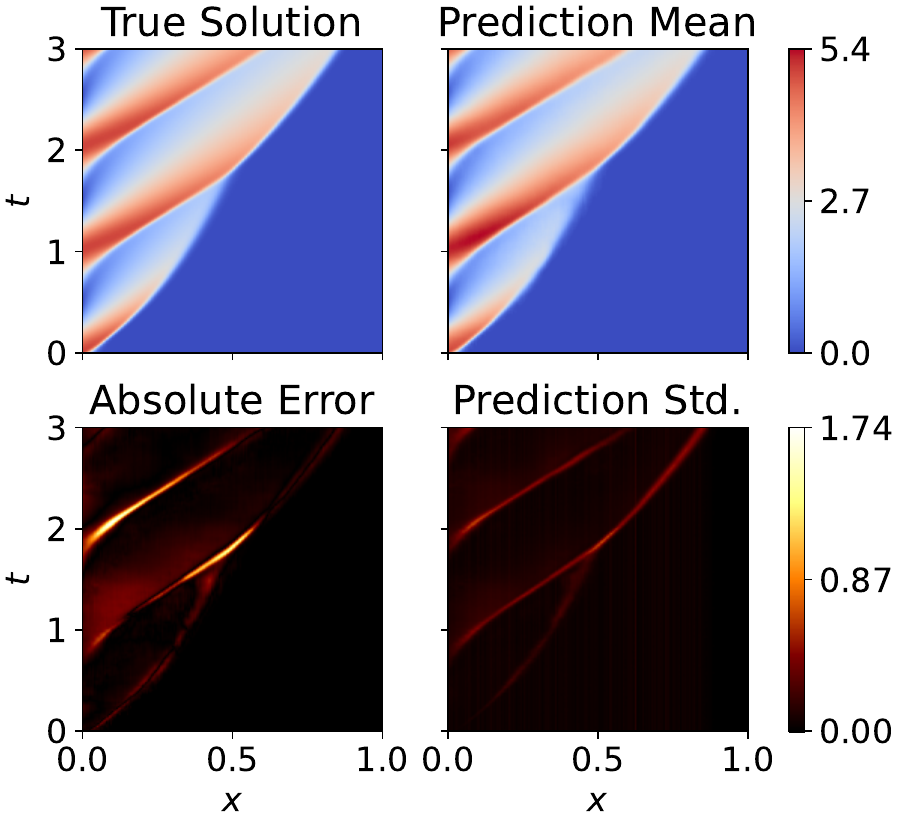}
			\captionof{figure}{Burgers' equation: solution, error, \& prediction statistics for a test trajectory. See movie S3.}
			\label{fig:burgers_state}
		\end{minipage}
		\vspace{-1em}
	\end{table}
	\begin{figure}[t]
		\centering
		\includegraphics[width=\linewidth]{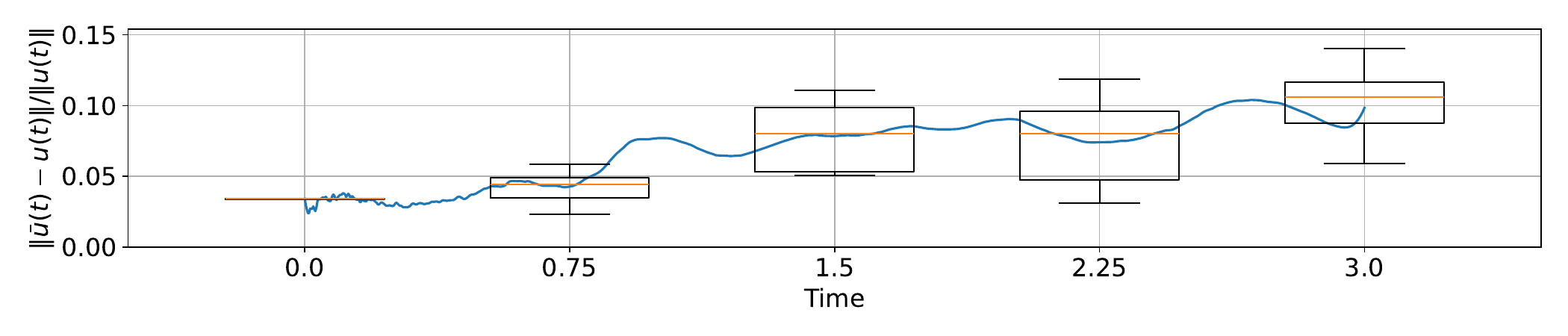}
		\caption{Burgers' equation: error distribution on test set over time for $d=5$.}
		\label{fig:burgers_error}
	\end{figure}
	
	\subsection*{Example 2: Forced, parametrized Burgers' equation}
	
	We next consider the viscous one-dimensional Burgers' equation over $x\in [0, 1]$ and $t\in [0, 3]$. This problem is parametrized by viscosity $\nu \in [0.05,0.1]$ and the time-dependent forcing is given as $f(t) = \alpha_1 \cos(2\pi \omega t)$, where $\omega \in [0.8, 1]$. The initial condition is $v(x, 0) = \alpha_1 \exp (-x^2/\alpha_2)$, where $\alpha_1 = 5$ and $\alpha_2 = 0.001$. The forcing is applied to the left boundary, i.e., $v(0, t) = f(t)$, and the right boundary has a zero Dirichlet condition. We discretize the governing equations using a finite-difference scheme on a spatial mesh with $D=500$ equispaced points and use a time-marching scheme with $\Delta t = 0.001$, resulting in $1001$ time samples in each FOM trajectory.
	
	We generate our dataset by running the FOM for different realizations of $\nu$ and $\omega$. The training set contains 100 trajectories, the validation set contains 10, and the test set contains 10. Applying the principle for estimating intrinsic manifold dimensionality discussed by Lee and Carlberg~\cite{LeeCarlberg2020}, we observe that any state snapshot in our problem is determined by viscosity $\nu$, forcing frequency $\omega$, and time $t$. This implies a manifold dimension of at least 3, which we use as a lower bound to explore latent dimensions $d \in \{3, 4, 5\}$; see Section~\ref{sec:burgersdetails} for additional details.
	
	The results are summarized in Table~\ref{tab:burgers_comparison}. For this more complex problem with variations in parameters and forcing, the SINDy-based approaches (POD-SINDy and AE-SINDy) struggle to achieve competitive accuracy compared to the neural differential equation models, with AE-SINDy showing some improvement for $d=4$. We observe considerable variation in PNODE/PNSDE prediction accuracy across different latent dimensions due to difficulties in converging to a model that generalizes well. In contrast, our proposed approach (PA) consistently delivers good accuracy, particularly for $d=4$ and $d=5$, while maintaining significantly shorter training times than PNODE/PNSDE. For instance, with $d=4$, PA achieves a mean error of $0.065$ with a training time of $3.281$ hours, outperforming PNODE ($\varepsilon_\mu=0.707$, $T_{\rm tr}=6.035$) and PNSDE ($\varepsilon_\mu=0.718$, $T_{\rm tr}=4.407$) in both metrics. Figures~\ref{fig:burgers_state} and~\ref{fig:burgers_error} show visualizations of our stochastic ROM predictions. The error distribution plot shows that as shock boundaries form, local error increases sharply. Notably, the uncertainty in the stochastic ROM prediction is strongly correlated with these high-error regions, increasing along the shock boundaries.
	
	\begin{table}[t]
		\centering
		\caption{Fluid flow with control: model comparison for varying ROM dim. $d$ over mean error $\varepsilon_\mu$, standard deviation of error $\varepsilon_\sigma$, and training time $T_{\rm tr}$ in hours. PA: present approach.}
		\label{tab:flowcontrol_comparison}
		\begin{tabular}{cc|ccccc}
			\toprule
			& & & POD- & AE- & & \\
			$d$ & Metric & PA & SINDy & SINDy & PNODE & PNSDE \\
			\midrule
			& $\varepsilon_\mu$ & $0.043$ & $0.088$ & $0.078$ & $0.100$ & $0.095$ \\
			$7$ & $\varepsilon_\sigma$ & $0.030$ & $0.026$ & $0.046$ & $0.030$ & $0.025$ \\
			& $T_{\rm tr}$ & $3.223$ & $0.105$ & $0.760$ & $25.92$ & $19.31$ \\
			\midrule
			& $\varepsilon_\mu$ & $0.046$ & $0.098$ & $0.078$ & $0.078$ & $0.103$ \\
			$8$ & $\varepsilon_\sigma$ & $0.020$ & $0.029$ & $0.039$ & $0.021$ & $0.032$ \\
			& $T_{\rm tr}$ & $3.221$ & $0.102$ & $0.771$ & $19.95$ & $19.83$ \\
			\midrule
			& $\varepsilon_\mu$ & $0.036$ & $0.095$ & $0.073$ & $0.076$ & $0.091$ \\
			$9$ & $\varepsilon_\sigma$ & $0.026$ & $0.028$ & $0.037$ & $0.021$ & $0.020$ \\
			& $T_{\rm tr}$ & $3.256$ & $0.100$ & $0.770$ & $23.53$ & $19.18$ \\
			\bottomrule
		\end{tabular}
	\end{table}
	
	\subsection*{Example 3: Fluid flow with control input}
	
	To demonstrate the scalability of our approach, we now consider two-dimensional fluid flow around a cylinder at $Re=100$ with time-dependent forcing, a test case adapted from the work by Rabault et al.~\cite{Rabault_2019_Flow_Control_DRL} on flow control of the Von K\'{a}rm\'{a}n vortex street using deep reinforcement learning. The forcing takes the form of jets on the top and bottom of the cylinder, whose flow rates $Q_1,\ Q_2$ are controlled. No net mass is injected into or removed from the system by the jets, so $f(t) = Q_1 = -Q_2$. We model the flow rate as a zero-mean Gaussian process with covariance function $k(t_1, t_2) = \alpha \exp{(-(t_1 - t_2)^2/(2 \ell^2))}$, 
	where $\ell = T_v$ and $\alpha = 10^{-5}$.
	
	The FOM in this case is a fluid flow simulation on an unstructured mesh~\cite{Rabault_2019_Flow_Control_DRL}. The velocity and pressure fields on the mesh are resampled onto a $440 \times 80$ uniform spatial grid, leading to a $D=105,600-$dimensional state. 
	We generate $120$ training, $10$ validation, and $10$ test trajectories for different realizations of the flow rate over a time-interval spanning $2$-times the vortex shedding period $T_v$. 
	We vary the latent state dimension $d\in \{7, 8, 9\}$ for this test case; additional details can be found in Section~\ref{sec:flowdetails}.
	
	The results are summarized in Table~\ref{tab:flowcontrol_comparison}. In this high-dimensional fluid flow problem, our approach consistently achieves the lowest mean prediction errors across all tested latent dimensions ($d=7,8,9$) compared to the baselines. For instance, with $d=9$, PA yields a mean error $\varepsilon_\mu$ of $0.036$, outperforming PNODE ($0.076$), PNSDE ($0.091$), AE-SINDy ($0.073$), and POD-SINDy ($0.095$). The training time of our approach ($T_{\rm tr} \approx 3.2$ hours) is substantially shorter than that of PNODE ($19-26$ hours) and PNSDE ($\approx19$ hours). This significant speed-up, coupled with high accuracy, underscores the computational efficiency and effectiveness of our solver-free, amortized training strategy for large-scale, complex dynamical systems. While SINDy-based methods train faster than PNODE/PNSDE, their accuracy remains notably lower than our approach for this problem.
	
\begin{figure}[t]
		\centering
		\parbox{0.5\textwidth}{\centering \large $x$-Velocity}%
		\parbox{0.5\textwidth}{\centering \large $y$-Velocity}%
		\vspace{0.5em}
		\includegraphics[width=0.5\textwidth]{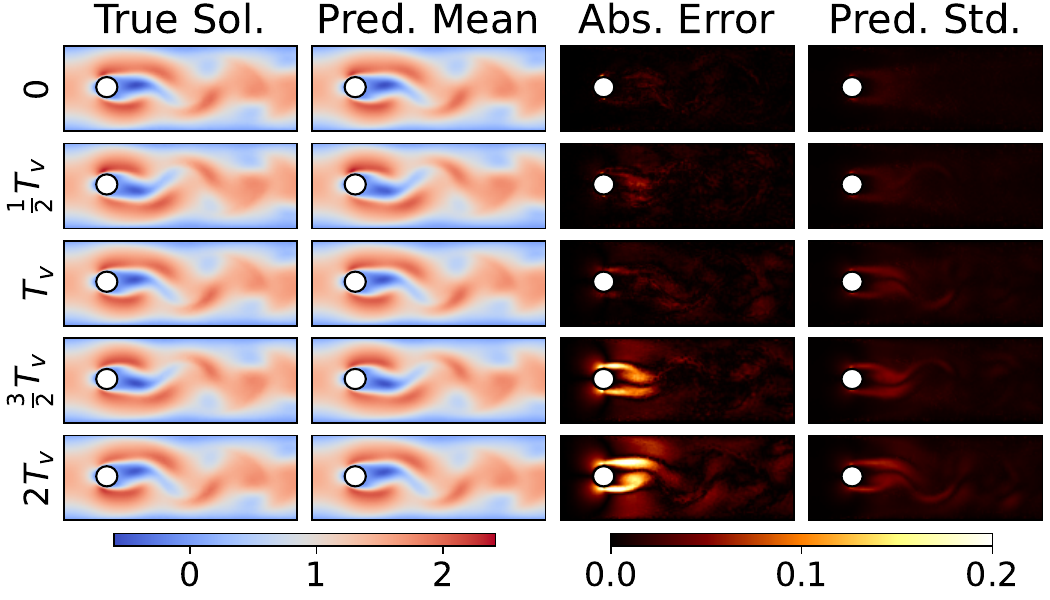}%
		\includegraphics[width=0.5\linewidth]{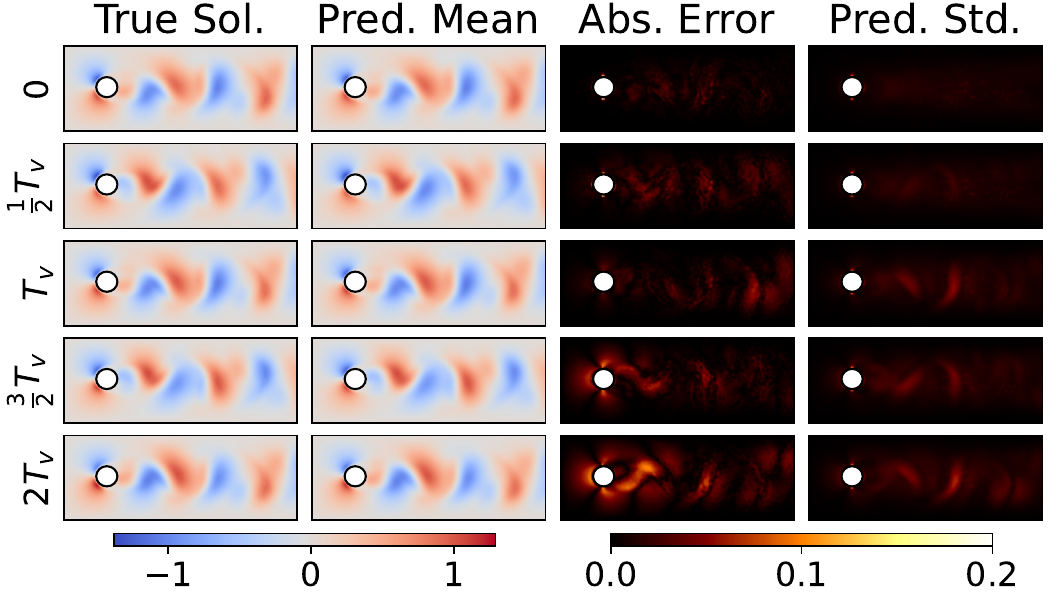}%
		\caption{Fluid flow with control: solution, error, \& prediction statistics for a test trajectory. Only the velocity field on the left half of the domain is shown. See movie S4.}\label{fig:flowcontrol_state}
	\end{figure}

\begin{figure}[t]
		\centering
		\includegraphics[width=\linewidth]{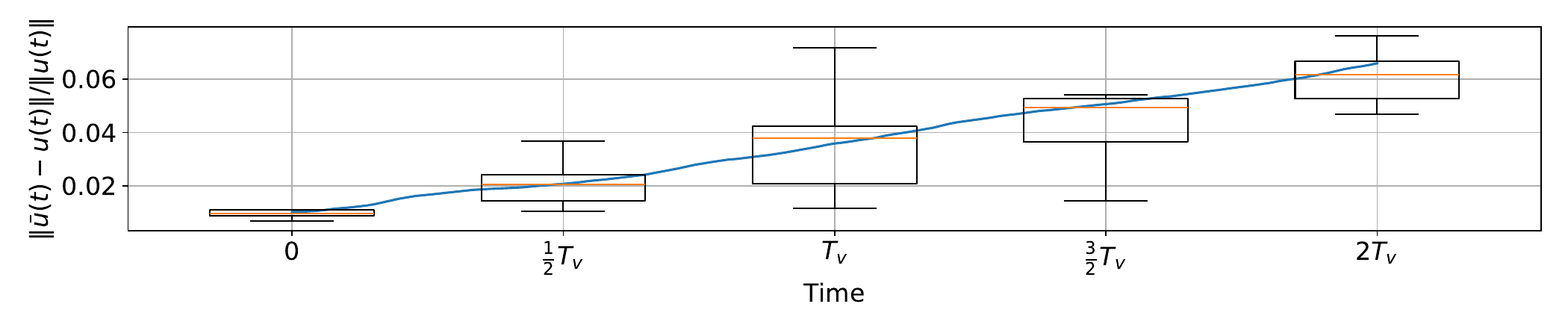}%
		\caption{Fluid flow with control: error distribution on test set over time for $d=9$.}\label{fig:flowcontrol_error}
	\end{figure}

	Visualizations of the predictions and error statistics of our stochastic ROM are provided in Figures~\ref{fig:flowcontrol_state} and~\ref{fig:flowcontrol_error}. Similar to the reaction-diffusion system, the mean error for our approach  grows approximately linearly over time, starting at $0.01$ and reaching around $0.06$ by the end of the simulation period. This is mainly due to the same reason discussed earlier: the ROM slowly drifts out of phase with the true
	solution. The prediction snapshots for a sample test trajectory show that the prediction uncertainty is initially uniform but becomes more pronounced in the cylinder wake as errors accumulate, aligning with regions of complex flow dynamics. See movie S4 for a visualization of the prediction error over time.
	
	\section{Discussion}
	
	In this work, we have developed a powerful new approach for constructing stochastic ROMs that captures the dependence of the dynamics on system parameters and forcing conditions while enabling joint inference of the ROM along with a probabilistic autoencoder. By eliminating forward solvers from the training process, 
	we achieve a significant breakthrough in computational efficiency and advance the state-of-the-art in data-driven ROM methodology. The amortized SVI framework decouples the training cost from the number of training trajectories, time-step resolution, and system stiffness. This enables efficient distillation of complex relationships between the dynamics, system parameters, and time-dependent forcing from large-scale datasets.
	
	
	Our approach offers unique flexibility in model construction. While we can learn parsimonious latent models comparable to those obtained using SINDy~\cite{Champion_2019_Data_Driven_Coordinates}, we do so without requiring time derivatives in the training data. This represents a significant practical advantage since numerical differentiation often amplifies measurement noise. Indeed, the reaction-diffusion test case -- where noise was added independently to state samples and their time derivatives -- represents an ideal scenario for SINDy's performance on noisy datasets.
	
	Even though our amortized construction of the latent variational distribution offers substantial computational advantages over existing approaches, the ELBO evaluation cost remains a limiting factor. Further computational efficiency gains would enable application to even more demanding problems, such as high-resolution atmospheric modeling~\cite{Yu_2024_ClimSim} and learning dynamical systems from large-scale neural activity datasets~\cite{Azabou_2023_Neural_Decoding}.
	
	Our implementation demonstrates only a subset of possible statistical priors for enhancing ROM performance. Future work could incorporate domain knowledge by encoding physical features -- such as dissipation or oscillation patterns -- into specialized parametrizations of the latent drift function. For systems with known FOMs, Galerkin projection could inform stronger priors on the latent SDE, potentially improving both training efficiency and model interpretability.

	The methodology presented here bridges the spectrum between physics-driven and data-driven ROM. Although our focus has been primarily on the data-driven end -- working with time-series data corresponding to variations in parameters and forcing without access to the FOM -- our framework's ability to incorporate known physics makes it adaptable across this entire spectrum. More specifically, the ability to incorporate full or partial statistical priors derived from structural or physical assumptions provides a natural bridge between purely data-driven and physics-informed modeling paradigms. This versatility also opens up exciting new possibilities for tackling challenging problems such as probabilistic coarse-grained modeling~\cite{Wang_2019_CG_MD}, learning Koopman representations~\cite{Brunton_2022_Koopman} of complex dynamical systems, and inferring models from indirect observations~\cite{Course_2023_State_Estimation, Raissi_2020_Learning_Fluids}, positioning our approach as a foundational tool for scientific machine learning.
	
	\paragraph{Data/code availability}
	A software library implementing the proposed stochastic ROM methodology along with training and evaluation datasets to reproduce our results is available on GitHub: \url{https://github.com/ailersic/rom-visde}.
	
	\paragraph{Acknowledgments}
	This research is supported by an NSERC Discovery Grant, a Queen Elizabeth II Graduate Scholarship in Science and Technology, and a grant from the Data Sciences Institute at the University of Toronto.
	
	\bibliography{SDELearning.bib}

@article{Blanchet_2020_Exact_Simulation_Multivariate,
	ISSN = {00018678, 14756064},
	URL = {https://www.jstor.org/stable/48654467},
	author = {Jose Blanchet and Fan Zhang},
	journal = {Advances in Applied Probability},
	number = {4},
	pages = {pp. 1003--1034},
	publisher = {Applied Probability Trust},
	title = {Exact Simulation for Multivariate {I}t\^o Diffusions},
	urldate = {2025-06-30},
	volume = {52},
	year = {2020}
}

@article{Beskos_2005_Exact_Simulation,
	author = {Beskos, Alexandros and Roberts, Gareth O.},
	year = {2005},
	pages = {2422--2444},
	title = {Exact simulation of diffusions},
	volume = {15},
	journal = {The Annals of Applied Probability},
	doi = {10.1214/105051605000000485}
}

@article{Elerian_1998_Milstein_Density,
	title={A note on the existence of a closed form conditional transition density for the {M}ilstein scheme},
	author={Elerian, Ola},
	journal={Economics discussion paper},
	pages={W18},
	year={1998},
	publisher={Nuffield College, University of Oxford}
}

@article{Uhlenbeck_1930_OU_Process,
	title = {On the Theory of the {B}rownian Motion},
	author = {Uhlenbeck, George E. and Ornstein, Leonard S.},
	journal = {Phys. Rev.},
	volume = {36},
	issue = {5},
	pages = {823--841},
	numpages = {0},
	year = {1930},
	month = {Sep},
	publisher = {American Physical Society},
	doi = {10.1103/PhysRev.36.823},
	url = {https://link.aps.org/doi/10.1103/PhysRev.36.823}
}

@article{Cox_1985_CIR_Model,
	ISSN = {00129682, 14680262},
	URL = {http://www.jstor.org/stable/1911242},
	author = {John C. Cox and Jonathan E. Ingersoll and Stephen A. Ross},
	journal = {Econometrica},
	number = {2},
	pages = {385--407},
	publisher = {[Wiley, Econometric Society]},
	title = {A Theory of the Term Structure of Interest Rates},
	urldate = {2025-06-30},
	volume = {53},
	year = {1985}
}

@article{Milstein_1979_Method_SDEs,
	author = {Milstein, Grigori N.},
	title = {A Method of Second-Order Accuracy Integration of Stochastic Differential Equations},
	journal = {Theory of Probability \& Its Applications},
	volume = {23},
	number = {2},
	pages = {396-401},
	year = {1979},
	doi = {10.1137/1123045},
	URL = {https://doi.org/10.1137/1123045},
	eprint = {https://doi.org/10.1137/1123045}
}

@article{Ait_Sahalia_2002_Method_SDEs,
	author = {A\"it-Sahalia, Yacine},
	title = {Maximum Likelihood Estimation of Discretely Sampled Diffusions: A Closed-form Approximation Approach},
	journal = {Econometrica},
	volume = {70},
	number = {1},
	pages = {223-262},
	doi = {https://doi.org/10.1111/1468-0262.00274},
	url = {https://onlinelibrary.wiley.com/doi/abs/10.1111/1468-0262.00274},
	eprint = {https://onlinelibrary.wiley.com/doi/pdf/10.1111/1468-0262.00274},
	year = {2002}
}

@article{Craigmile_2023_Inference_SDEs_Review,
	author = {Craigmile, Peter and Herbei, Radu and Liu, Ge and Schneider, Grant},
	title = {Statistical inference for stochastic differential equations},
	journal = {WIREs Computational Statistics},
	volume = {15},
	number = {2},
	pages = {e1585},
	doi = {https://doi.org/10.1002/wics.1585},
	url = {https://wires.onlinelibrary.wiley.com/doi/abs/10.1002/wics.1585},
	eprint = {https://wires.onlinelibrary.wiley.com/doi/pdf/10.1002/wics.1585},
	year = {2023}
}

@inproceedings{Oh_2024_Stable_NSDEs,
	title={Stable Neural Stochastic Differential Equations in Analyzing Irregular Time Series Data},
	author={Oh, YongKyung and Lim, Dongyoung and Kim, Sungil},
	booktitle={The Twelfth International Conference on Learning Representations},
	year={2024}
}

@article{Rubanova_2019_NODE,
	author = {Rubanova, Yulia and Chen, Ricky T. Q. and Duvenaud, David K},
	journal = {Advances in Neural Information Processing Systems},
	pages = {},
	publisher = {Curran Associates, Inc.},
	title = {Latent Ordinary Differential Equations for Irregularly-Sampled Time Series},
	volume = {32},
	year = {2019}
}

@article{De_Silva_2020_PySINDy,
	doi = {10.21105/joss.02104},
	url = {https://doi.org/10.21105/joss.02104},
	year = {2020},
	publisher = {The Open Journal},
	volume = {5},
	number = {49},
	pages = {2104},
	author = {Brian de Silva and Kathleen Champion and Markus Quade and Jean-Christophe Loiseau and J. Kutz and Steven Brunton},
	title = {PySINDy: A Python package for the sparse identification of nonlinear dynamical systems from data},
	journal = {Journal of Open Source Software}
}

@article{Kaptanoglu_2022_PySINDy,
	doi = {10.21105/joss.03994},
	url = {https://doi.org/10.21105/joss.03994},
	year = {2022},
	publisher = {The Open Journal},
	volume = {7},
	number = {69},
	pages = {3994},
	author = {Alan A. Kaptanoglu and Brian M. de Silva and Urban Fasel and Kadierdan Kaheman and Andy J. Goldschmidt and Jared Callaham and Charles B. Delahunt and Zachary G. Nicolaou and Kathleen Champion and Jean-Christophe Loiseau and J. Nathan Kutz and Steven L. Brunton},
	title = {PySINDy: A comprehensive Python package for robust sparse system identification},
	journal = {Journal of Open Source Software}
}

@article{Conti_2023_ROM_Parametrized_SINDy,
	title = {Reduced order modeling of parametrized systems through autoencoders and SINDy approach: continuation of periodic solutions},
	journal = {Computer Methods in Applied Mechanics and Engineering},
	volume = {411},
	pages = {116072},
	year = {2023},
	issn = {0045-7825},
	doi = {https://doi.org/10.1016/j.cma.2023.116072},
	url = {https://www.sciencedirect.com/science/article/pii/S0045782523001962},
	author = {Paolo Conti and Giorgio Gobat and Stefania Fresca and Andrea Manzoni and Attilio Frangi}
}

@article{Nicolaou_2023_Data_Driven_Parametrized,
	title = {Data-driven discovery and extrapolation of parameterized pattern-forming dynamics},
	author = {Nicolaou, Zachary G. and Huo, Guanyu and Chen, Yihui and Brunton, Steven L. and Kutz, J. Nathan},
	journal = {Phys. Rev. Res.},
	volume = {5},
	issue = {4},
	pages = {L042017},
	numpages = {7},
	year = {2023},
	month = {Nov},
	publisher = {American Physical Society},
	doi = {10.1103/PhysRevResearch.5.L042017},
	url = {https://link.aps.org/doi/10.1103/PhysRevResearch.5.L042017}
}

@book{sarkka2019book,
	title={Applied stochastic differential equations},
	author={S{\"a}rkk{\"a}, Simo and Solin, Arno},
	volume={10},
	series={Institute of Mathematical Statistics Textbooks},
	year={2019},
	publisher={Cambridge University Press},
	address={Cambridge}
}

@article{Baptista_2022_Gradient_Reduction,
	title={Gradient-based data and parameter dimension reduction for {B}ayesian models: an information theoretic perspective}, 
	author={Ricardo Baptista and Youssef Marzouk and Olivier Zahm},
	year={2022},
	eprint={2207.08670},
	archivePrefix={arXiv},
	primaryClass={stat.CO},
	url={https://arxiv.org/abs/2207.08670},
	journal={arXiv preprint}
}

@article{Romor2023,
	author = {Romor, Francesco and Stabile, Giovanni and Rozza, Gianluigi},
	journal = {Journal of Scientific Computing},
	number = {3},
	title = {Nonlinear Manifold Reduced-Order Models with Convolutional Autoencoders and Reduced Over-Collocation Method},
	volume = {94},
	year = {2023}
}

@article{Wang_2019_CG_MD,
	author={Wang, Jiang
	and Olsson, Simon
	and Wehmeyer, Christoph
	and P{\'e}rez, Adri{\`a}
	and Charron, Nicholas E.
	and de Fabritiis, Gianni
	and No{\'e}, Frank
	and Clementi, Cecilia},
	title={Machine Learning of Coarse-Grained Molecular Dynamics Force Fields},
	journal={ACS Central Science},
	year={2019},
	month={May},
	day={22},
	publisher={American Chemical Society},
	volume={5},
	number={5},
	pages={755-767},
	issn={2374-7943}
}

@article{Aka_2025_Balanced_Neural_ODEs,
	title={Balanced Neural {ODE}s: nonlinear model order reduction and {K}oopman operator approximations}, 
	author={Julius Aka and Johannes Brunnemann and Jörg Eiden and Arne Speerforck and Lars Mikelsons},
	year={2025},
	eprint={2410.10174},
	archivePrefix={arXiv},
	primaryClass={cs.LG},
	journal={arXiv preprint}
}

@article{Wan_2018_ROM,
	author = {Wan, Zhong Yi AND Vlachas, Pantelis AND Koumoutsakos, Petros AND Sapsis, Themistoklis},
	journal = {PLOS ONE},
	publisher = {Public Library of Science},
	title = {Data-assisted reduced-order modeling of extreme events in complex dynamical systems},
	year = {2018},
	month = {05},
	volume = {13},
	pages = {1-22},
	number = {5},
	
}

@article{Raissi_2020_Learning_Fluids,
	author = {Maziar Raissi  and Alireza Yazdani  and George Em Karniadakis },
	title = {Hidden fluid mechanics: Learning velocity and pressure fields from flow visualizations},
	journal = {Science},
	volume = {367},
	number = {6481},
	pages = {1026-1030},
	year = {2020}
}

@article{Brunton_2022_Koopman,
	author = {Brunton, Steven L. and Budi\v{s}i\'{c}, Marko and Kaiser, Eurika and Kutz, J. Nathan},
	title = {Modern {K}oopman Theory for Dynamical Systems},
	journal = {SIAM Review},
	volume = {64},
	number = {2},
	pages = {229-340},
	year = {2022}
}

@article{Benner2015,
	author = {P. Benner and S. Gugercin and K. Willcox},
	journal = {SIAM Review},
	number = {4},
	title = {A survey of projection-based model reduction methods for parametric dynamical systems},
	volume = {57},
	year = {2015}
}

@article{Barnett2022,
	title = {Quadratic approximation manifold for mitigating the {K}olmogorov barrier in nonlinear projection-based model order reduction},
	journal = {Journal of Computational Physics},
	volume = {464},
	pages = {111348},
	year = {2022},
	author = {Joshua Barnett and Charbel Farhat},
}

@article{Audouze2009,
	author = {Audouze, Christophe and De Vuyst, Florian and Nair, Prasanth B.},
	title = {Reduced-order modeling of parameterized {PDE}s using time–space-parameter principal component analysis},
	journal = {International Journal for Numerical Methods in Engineering},
	volume = {80},
	number = {8},
	pages = {1025-1057},
	year = {2009}
}

@article{Audouze2013,
	author = {Audouze, Christophe and De Vuyst, Florian and Nair, Prasanth B.},
	title = {Nonintrusive reduced-order modeling of parametrized time-dependent partial differential equations},
	journal = {Numerical Methods for Partial Differential Equations},
	volume = {29},
	number = {5},
	pages = {1587-1628},
	year = {2013}
}

@article{Xing2015,
	author = {Xing, Wei W. and Shah, Akeel A. and Nair, Prasanth B.},
	title = {Reduced dimensional {G}aussian process emulators of parametrized partial differential equations based on Isomap},
	journal = {Proceedings of the Royal Society A: Mathematical, Physical and Engineering Sciences},
	volume = {471},
	number = {20140697},
	year = {2015}
}

@article{Xing2016,
	author = {Xing, Wei W. and Triantafyllidis, Vasileios and Shah, Akeel A. and Nair, Prasanth B. and Zabaras, Nicholas},
	title = {Manifold learning for the emulation of spatial fields from computational models},
	journal = {Journal of Computational Physics},
	volume = {326},
	pages = {666-690},
	year = {2016},
}

@article{DuanHesthaven2024,
	title = {Non-intrusive data-driven reduced-order modeling for time-dependent parametrized problems},
	journal = {Journal of Computational Physics},
	volume = {497},
	pages = {112621},
	year = {2024},
	author = {Junming Duan and Jan S. Hesthaven}
}

@article{XuDuraisamy2020,
	author = {Jiayang Xu and Karthik Duraisamy},
	title = {Multi-level convolutional autoencoder networks for parametric prediction of spatio-temporal dynamics},
	journal = {Computer Methods in Applied Mechanics and Engineering},
	volume = {372},
	pages = {113379},
	year = {2020}
}

@article{Maulik2021,
	author = {Romit Maulik and Bethany Lusch and Prasanna Balaprakash},
	title = {Reduced-order modeling of advection-dominated systems with recurrent neural networks and convolutional autoencoders},
	journal = {Physics of Fluids},
	volume = {33},
	number = {3},
	pages = {037106},
	year = {2021}
}

@article{Peherstorfer2022,
	author = {Peherstorfer, Benjamin},
	title = {Breaking the {K}olmogorov barrier with nonlinear model reduction},
	journal = {Notices of the American Mathematical Society},
	volume = {69},
	number = {5},
	pages = {725-733},
	year = {2022}
}

@inproceedings{mildenhall2020nerf,
 title={NeRF: Representing Scenes as Neural Radiance Fields for View Synthesis},
 author={Ben Mildenhall and Pratul P. Srinivasan and Matthew Tancik and Jonathan T. Barron and Ravi Ramamoorthi and Ren Ng},
 year={2020},
 booktitle={European Conference on Computer Vision (ECCV)}
}

@article{tancik2020fourfeat,
    title={Fourier Features Let Networks Learn High Frequency Functions in Low Dimensional Domains},
    author={Matthew Tancik and Pratul P. Srinivasan and Ben Mildenhall and Sara Fridovich-Keil and Nithin Raghavan and Utkarsh Singhal and Ravi Ramamoorthi and Jonathan T. Barron and Ren Ng},
    journal={NeurIPS},
    year={2020}
}

@article{Kingma_2017_Adam,
	title={Adam: A Method for Stochastic Optimization}, 
	author={Diederik P. Kingma and Jimmy Ba},
	year={2017},
	eprint={1412.6980},
	archivePrefix={arXiv},
	primaryClass={cs.LG},
	url={https://arxiv.org/abs/1412.6980},
	journal={arXiv preprint}
}

@article{Yu_2024_ClimSim,
	title={{ClimSim-Online:} A Large Multi-scale Dataset and Framework for Hybrid {ML}-physics Climate Emulation}, 
	author={Sungduk Yu and Zeyuan Hu and Akshay Subramaniam and Walter Hannah and Liran Peng and Jerry Lin and Mohamed Aziz Bhouri and Ritwik Gupta and Björn Lütjens and Justus C. Will and Gunnar Behrens and Julius J. M. Busecke and Nora Loose and Charles I. Stern and Tom Beucler and Bryce Harrop and Helge Heuer and Benjamin R. Hillman and Andrea Jenney and Nana Liu and Alistair White and Tian Zheng and Zhiming Kuang and Fiaz Ahmed and Elizabeth Barnes and Noah D. Brenowitz and Christopher Bretherton and Veronika Eyring and Savannah Ferretti and Nicholas Lutsko and Pierre Gentine and Stephan Mandt and J. David Neelin and Rose Yu and Laure Zanna and Nathan Urban and Janni Yuval and Ryan Abernathey and Pierre Baldi and Wayne Chuang and Yu Huang and Fernando Iglesias-Suarez and Sanket Jantre and Po-Lun Ma and Sara Shamekh and Guang Zhang and Michael Pritchard},
	year={2024},
	eprint={2306.08754},
	archivePrefix={arXiv},
	primaryClass={cs.LG},
	url={https://arxiv.org/abs/2306.08754},
	journal={arXiv preprint}
}

@article{
	Azabou_2023_Neural_Decoding,
	title={A Unified, Scalable Framework for Neural Population Decoding},
	author={Mehdi Azabou and Vinam Arora and Venkataramana Ganesh and Ximeng Mao and Santosh Nachimuthu and Michael Mendelson and Blake Richards and Matthew Perich and Guillaume Lajoie and Eva L. Dyer},
	journal={Advances in Neural Information Processing Systems},
	year={2023},
}

@article{Kim_2022_PINN_ROM,
	title = {A fast and accurate physics-informed neural network reduced order model with shallow masked autoencoder},
	journal = {Journal of Computational Physics},
	volume = {451},
	pages = {110841},
	year = {2022},
	issn = {0021-9991},
	url = {https://www.sciencedirect.com/science/article/pii/S0021999121007361},
	author = {Youngkyu Kim and Youngsoo Choi and David Widemann and Tarek Zohdi}
}

@article{Huang_2023_ROM_Chaotic_Problems,
	title = {Predictive reduced order modeling of chaotic multi-scale problems using adaptively sampled projections},
	journal = {Journal of Computational Physics},
	volume = {491},
	pages = {112356},
	year = {2023},
	issn = {0021-9991},
	author = {Cheng Huang and Karthik Duraisamy}
}

@article{Champion_2019_Data_Driven_Coordinates,
	author = {Kathleen Champion  and Bethany Lusch  and J. Nathan Kutz  and Steven L. Brunton },
	title = {Data-driven discovery of coordinates and governing equations},
	journal = {Proceedings of the National Academy of Sciences},
	volume = {116},
	number = {45},
	pages = {22445-22451},
	year = {2019}
}

@article{Kingma_2022_Autoencoding_Variational_Bayes,
	title={Auto-Encoding Variational {B}ayes}, 
	author={Diederik P. Kingma and Max Welling},
	year={2022},
	eprint={1312.6114},
	archivePrefix={arXiv},
	primaryClass={stat.ML},
	url={https://arxiv.org/abs/1312.6114},
	journal={arXiv preprint}
}

@misc{Falcon_2019_PyTorch_Lightning,
	author = {Falcon, William and {The PyTorch Lightning team}},
	license = {Apache-2.0},
	month = mar,
	title = {{PyTorch Lightning}},
	version = {1.4},
	year = {2019},
	publisher = {GitHub},
	journal = {GitHub repository},
	howpublished = {\url{https://github.com/rtqichen/torchdiffeq}},
}

@inproceedings{Chen_2018_NODEs,
	author = {Chen, Ricky T. Q. and Rubanova, Yulia and Bettencourt, Jesse and Duvenaud, David},
	title = {Neural ordinary differential equations},
	year = {2018},
	publisher = {Curran Associates Inc.},
	address = {Red Hook, NY, USA},
	booktitle = {Proceedings of the 32nd International Conference on Neural Information Processing Systems},
	pages = {6572–6583},
	numpages = {12},
	location = {Montr\'{e}al, Canada},
	series = {NIPS'18}
}

@misc{Chen_2018_torchdiffeq,
	author={Chen, Ricky T. Q.},
	title={torchdiffeq: {PyTorch} Implementation of Differentiable {ODE} Solvers},
	year={2018},
	publisher = {GitHub},
	journal = {GitHub repository},
	howpublished = {\url{https://github.com/rtqichen/torchdiffeq}},
}

@book{Logg_2011_FEniCS,
	author = {Logg, Anders and Wells, Garth and Mardal, Kent-Andre},
	year = {2011},
	month = {April},
	pages = {},
	title = {Automated Solution of Differential Equations by the Finite Element Method: The FEniCS Book},
	volume = {84},
	isbn = {978-3-642-23098-1},
	series = {Lecture Notes in Computational Science and Engineering},
	publisher = {Springer},
	address = {Berlin, Heidelberg}
}

@article{Fries_2022_LaSDI,
	title = {{LaSDI}: Parametric Latent Space Dynamics Identification},
	journal = {Computer Methods in Applied Mechanics and Engineering},
	volume = {399},
	pages = {115436},
	year = {2022},
	issn = {0045-7825},
	author = {William D. Fries and Xiaolong He and Youngsoo Choi}
}

@article{Rabault_2019_Flow_Control_DRL,
	title={Artificial neural networks trained through deep reinforcement learning discover control strategies for active flow control},
	volume={865},
	journal={Journal of Fluid Mechanics},
	author={Rabault, Jean and Kuchta, Miroslav and Jensen, Atle and Réglade, Ulysse and Cerardi, Nicolas},
	year={2019},
	pages={281–302}
}

@article{Kidger_2022_Neural_Differential_Equations,
	title={On Neural Differential Equations}, 
	author={Patrick Kidger},
	year={2022},
	eprint={2202.02435},
	archivePrefix={arXiv},
	primaryClass={cs.LG},
	school={University of Oxford},
	journal={arXiv preprint}
}

@article{Lee_2021_Parameterized_Neural_ODE_ROM,
	author = {Lee, Kookjin  and Parish, Eric J. },
	title = {Parameterized neural ordinary differential equations: applications to computational physics problems},
	journal = {Proceedings of the Royal Society A: Mathematical, Physical and Engineering Sciences},
	volume = {477},
	number = {2253},
	pages = {20210162},
	year = {2021}
}

@article{Peherstorfer_2016_Operator_Inference,
	title={Data-driven operator inference for nonintrusive projection-based model reduction},
	author={Peherstorfer, Benjamin and Willcox, Karen},
	journal={Computer Methods in Applied Mechanics and Engineering},
	volume={306},
	pages={196--215},
	year={2016},
	publisher={Elsevier}
}

@article{Gillespie_2007_Chemical_Kinetics,
	author = "Gillespie, Daniel T.",
	title = "Stochastic Simulation of Chemical Kinetics", 
	journal= "Annual Review of Physical Chemistry",
	year = "2007",
	volume = "58",
	number = "Volume 58, 2007",
	pages = "35-55",
	publisher = "Annual Reviews",
	issn = "1545-1593",
	type = "Journal Article"
}

@Article{Rao_2002_Intracellular_Noise,
	author={Rao, Christopher V.
	and Wolf, Denise M.
	and Arkin, Adam P.},
	title={Control, exploitation and tolerance of intracellular noise},
	journal={Nature},
	year={2002},
	month={Nov},
	day={01},
	volume={420},
	number={6912},
	pages={231-237},
	issn={1476-4687}
}

@Book{Karniadakis_2005_Microflows,
	author={Karniadakis, George
	and Beskok, Ali
	and Aluru, Narayan},
	title={Microflows and Nanoflows Fundamentals and Simulation},
	series={Interdisciplinary Applied Mathematics, 29},
	year={2005},
	edition={1st},
	publisher={Springer},
	address={New York, NY},
	isbn={1-280-46097-0}
}

@article{Dupont_2019_ANODE,
	author = {Dupont, Emilien and Doucet, Arnaud and Teh, Yee Whye},
	journal = {Advances in Neural Information Processing Systems},
	editor = {H. Wallach and H. Larochelle and A. Beygelzimer and F. d\textquotesingle Alch\'{e}-Buc and E. Fox and R. Garnett},
	pages = {},
	publisher = {Curran Associates, Inc.},
	title = {Augmented Neural {ODE}s},
	volume = {32},
	year = {2019}
}

@article{Course_2023_Amortized_Reparameterization,
	author = {Course, Kevin and Nair, Prasanth B.},
	journal = {Advances in Neural Information Processing Systems},
	editor = {A. Oh and T. Naumann and A. Globerson and K. Saenko and M. Hardt and S. Levine},
	pages = {78296--78318},
	publisher = {Curran Associates, Inc.},
	title = {Amortized Reparametrization: Efficient and Scalable Variational Inference for Latent {SDE}s},
	volume = {36},
	year = {2023}
}

@Article{Course_2023_State_Estimation,
	author={Course, Kevin
	and Nair, Prasanth B.},
	title={State estimation of a physical system with unknown governing equations},
	journal={Nature},
	year={2023},
	month={Oct},
	day={01},
	volume={622},
	number={7982},
	pages={261-267},
	issn={1476-4687}
}

@article{Li_2020_Scalable_Gradients_VI_SDE,
	title = 	 {Scalable Gradients and Variational Inference for
	Stochastic Differential Equations },
	author =       {Li, Xuechen and Wong, Ting-Kam Leonard and Chen, Ricky T. Q. and Duvenaud, David K.},
	journal = 	 {Proceedings of The 2nd Symposium on
	Advances in Approximate Bayesian Inference},
	pages = 	 {1--28},
	year = 	 {2020},
	editor = 	 {Zhang, Cheng and Ruiz, Francisco and Bui, Thang and Dieng, Adji Bousso and Liang, Dawen},
	volume = 	 {118},
	series = 	 {Proceedings of Machine Learning Research},
	month = 	 {08 Dec},
	publisher =    {PMLR}
}

@article{Archambeau_2007_Variational_Inference_Diffusion,
	author = {Archambeau, C\'{e}dric and Opper, Manfred and Shen, Yuan and Cornford, Dan and Shawe-taylor, John},
	journal = {Advances in Neural Information Processing Systems},
	editor = {J. Platt and D. Koller and Y. Singer and S. Roweis},
	pages = {},
	publisher = {Curran Associates, Inc.},
	title = {Variational Inference for Diffusion Processes},
	volume = {20},
	year = {2007}
}

@article{Archambeau_2007_GP_Approx_SDE,
	title = 	 {Gaussian Process Approximations of Stochastic Differential Equations},
	author = 	 {Archambeau, C\'{e}dric and Cornford, Dan and Opper, Manfred and Shawe-Taylor, John},
	journal = 	 {Gaussian Processes in Practice},
	pages = 	 {1--16},
	year = 	 {2007},
	editor = 	 {Lawrence, Neil D. and Schwaighofer, Anton and Quiñonero Candela, Joaquin},
	volume = 	 {1},
	series = 	 {Proceedings of Machine Learning Research},
	address = 	 {Bletchley Park, UK},
	month = 	 {12--13 Jun},
	publisher =    {PMLR}
}

@article{FoxMiura1971,
author = {Fox, Richard L. and Miura, Hirokazu},
title = {An approximate analysis technique for design calculations},
journal = {AIAA Journal},
volume = {9},
number = {1},
pages = {177--179},
year = {1971}
}

@article{LeeCarlberg2020,
	Author = {Kookjin Lee and Kevin T. Carlberg},
	Journal = {Journal of Computational Physics},
	Pages = {108973},
	Title = {Model reduction of dynamical systems on nonlinear manifolds using deep convolutional autoencoders},
	Volume = {404},
	Year = {2020}}

@article{McQuarrie2023,
author = {McQuarrie, Shane A. and Khodabakhshi, Parisa and Willcox, Karen E.},
title = {Nonintrusive Reduced-Order Models for Parametric Partial Differential Equations via Data-Driven Operator Inference},
journal = {SIAM Journal on Scientific Computing},
volume = {45},
number = {4},
pages = {A1917-A1946},
year = {2023}
}

@book{Benner2022,
  author = {Benner, Peter and Schilders, Wil and Grivet-Talocia, Stefano and Quarteroni, Alfio and Rozza, Gianluigi and Silveira, Lu{\'\i}s Miguel},
  publisher = {De Gruyter},
  title = {Model order reduction},
  volume = {1-3},
  year = {2022},
  address = {Berlin}
}

	\newpage
	\appendix

	\section{PNODE/PNSDE details} \label{sec:pnodepnsde}
	
	In this section, we summarize the PNODE and PNSDE baseline models used in this work.

	\paragraph{PNODE formulation}
	
	In the original PNODE formulation of Lee and Parish~\cite{Lee_2021_Parameterized_Neural_ODE_ROM}, the forward pass proceeds as follows. For the $i$th trajectory with parameters $\mu_i$ and initial condition $u_i(0)$: (1)~the latent initial condition is obtained from a deterministic encoder $z_i(0) = h_\theta^{\rm enc}(u_i(0))$; (2) the parametrized neural ODE
	$z_i(t) = z_i(0) + \int_0^t \psi_\theta(z, t; \mu_i)\,{\rm d}t$
	is solved forward in time for a drift function $\psi_\theta$; and (3) the latent solution is mapped through a deterministic decoder to obtain the prediction $\overline{u}_i(t) = h_\theta^{\rm dec}(z_i(t))$. The parameters $\theta$ (encoder, decoder, and drift) are originally learned by minimizing the mean-squared error
	\begin{equation*}
		\mathcal{L}(\theta) = \frac{1}{N_T}\sum_{i=1}^{N_T}\frac{1}{N_i}\sum_{j=1}^{N_i}\left(u_{i,j} - \overline{u}_{i}(t_j)\right)^2.
	\end{equation*}
	
	In our numerical experiments, this MSE loss was not sufficient for large multi-trajectory datasets, as it does not impose a prior over the latent state. We therefore replace the MSE loss with an ELBO using a prior $p(z)$ on the latent state, as in variational autoencoders~\cite{Kingma_2022_Autoencoding_Variational_Bayes}, which has previously been used for training latent neural ODE models~\cite{Chen_2018_torchdiffeq, Rubanova_2019_NODE}:
	\begin{equation*}
		\mathcal{L}(\theta) = \sum_{i=1}^{N_T} \sum_{j=1}^{N_i} 
		\mathbb{E}_{ z_{i,j} }
		\left[ \log p_{\theta}^{\rm dec} \bigl( u_{i}(t_{i,j};  \mu_i) \mid z_{i,j} \bigr) \right]
		- \sum_{i=1}^{N_T} D_{\textrm{KL}}\bigl(p^{\rm enc}_\theta\bigl(z_{i}(0)\mid\mu_i\bigr) \,\|\, p(z)\bigr),
	\end{equation*}
	where $p_\theta^{\rm enc}(z_i(0)\mid \mu_i)$ is the initial-condition encoder, $p(z) = \mathcal{N}(0, I)$, and
	\begin{equation*}
		z_{i,j} = z_i(0) + \int_0^{t_{i,j}}\psi_\theta(z, t; \mu_i)\,{\rm d}t,
		\qquad 
		z_i(0) \sim p_\theta^{\rm enc}(z_i(0)\mid \mu_i).
	\end{equation*}

	Using an ELBO requires probabilistic encoders and decoders rather than deterministic ones, so we adopt probabilistic versions of these components. The encoder, decoder, and drift share the same neural architectures as the corresponding modules in our proposed approach for each example.
	
	In our numerical experiments, the PNODE model trained by maximizing the ELBO yields better generalization performance. For example, on the Burgers' equation dataset (see Example 2 in Section~\ref{sec:examples}) with ROM dimension $d=3$, the PNODE model attains a mean test error of $0.075$ with the ELBO, compared with $0.135$ when trained with the MSE loss.

	\paragraph{PNSDE formulation}
	Neural SDEs form a broad class of dynamical models~\cite{Oh_2024_Stable_NSDEs}. Our PNSDE baseline is obtained as a straightforward extension of the PNODE formulation. Retaining the same ELBO and model architecture, we augment the latent dynamics with a dispersion term:
	\begin{equation} \label{eqn:latentsde}
		z_{i,j} \sim z_i(0) + \int_0^{t_{i,j}}\psi_\theta(z, t; \mu_i)\,{\rm d}t 
		+ \int_0^{t_{i,j}}\Psi_\theta(t; \mu_i)\,{\rm d}\beta,
		\qquad 
		z_i(0) \sim p_\theta^{\rm enc}(z_i(0)\mid \mu_i),
	\end{equation}
	which is almost identical to the SDE prior in~\eqref{eqn:romsde}, except that it omits time-dependent forcing. Because the dynamics depend on parameters $\mu$, we refer to this baseline as a parametrized neural SDE (PNSDE) model.

	Parametric inference of $\theta$ in this formulation introduces a theoretical complication. Evaluating the log-likelihood term in the ELBO requires the distribution of the latent state $z$, induced by the stochastic process in~\eqref{eqn:latentsde}. The associated transition density $q(z(t)\mid t, z(0))$, which satisfies the Fokker–Planck equation, typically does not admit a closed form for general SDE systems, so the log-likelihood cannot be evaluated exactly.\footnote{This limitation does not arise in the SVI-based approach proposed in this work, despite its use of a latent SDE model, because the latent variational distribution over $z$ is parametrized as a linear Markov Gaussian process, rather than as a realization of an arbitrary stochastic process.} There is extensive literature on approximations of the transition density and on special cases with closed-form solutions; see Craigmile et al.~\cite{Craigmile_2023_Inference_SDEs_Review} for a recent overview.
	
	For the PNSDE baseline, we adopt an approximate transition density obtained from the Euler–Maruyama time-marching scheme,
	\begin{equation*}
		z(t+\Delta t) = z(t) + \psi_\theta(z(t), t; \mu)\,\Delta t + \Psi_\theta(t; \mu)\,\Delta \beta,
	\end{equation*}
	where $\Delta t$ is the time step and $\Delta \beta \sim \mathcal{N}(0, \Delta t I)$. Given $z(t)$ at time $t$, this yields a Gaussian stepwise transition density for $z(t+\Delta t)$,
	\begin{equation*}
		q\bigl(z(t+\Delta t)\mid \Delta t, z(t)\bigr) 
		= \mathcal{N}\bigl(z(t) + \psi_\theta(z(t), t; \mu)\,\Delta t,\,
		\Psi_\theta(t; \mu)\Psi_\theta^{\mathsf{T}}(t; \mu)\,\Delta t\bigr),
	\end{equation*}
	which we then use to approximate the full-trajectory transition density over $[0, t]$. Given an initial condition $z(0)$, a final time $t$, and time step $\Delta t$ such that $t = N \Delta t$ for $N \in \mathbb{Z}$,
	\begin{equation*}
		q(z(t)\mid t, z(0)) = \prod_{j=1}^{N} q\bigl(z(t_{j})\mid \Delta t, z(t_{j-1})\bigr),
		\quad \text{where } t_j = j\Delta t,\ t_0 = 0,
	\end{equation*}
	which serves as our transition-density approximation.
	
	Other time-marching schemes admit analogous approximate closed-form transition densities, including the Milstein method~\cite{Milstein_1979_Method_SDEs, Elerian_1998_Milstein_Density} and the A\"it-Sahalia method~\cite{Ait_Sahalia_2002_Method_SDEs}. Such approximations are, in practice, the only viable option for obtaining closed-form transition densities while retaining a flexible neural parametrization of the SDE. One may instead restrict attention to SDE families with known closed-form transition densities, such as the Cox–Ingersoll–Ross model~\cite{Cox_1985_CIR_Model} or the Ornstein–Uhlenbeck process~\cite{Uhlenbeck_1930_OU_Process}, but these do not admit neural parametrizations. A more general alternative is exact simulation of SDEs via rejection sampling, as in the methods of Beskos and Roberts~\cite{Beskos_2005_Exact_Simulation} for univariate systems and Blanchet and Zhang~\cite{Blanchet_2020_Exact_Simulation_Multivariate} for multivariate systems, but their computational cost is prohibitive (the expected termination time is infinite in the latter), and they are therefore not practical for tackling the class of problems considered in the present work.
	
	\section{Summary of parametrized terms} \label{sec:modules}
	We summarize below all components of the proposed stochastic ROM framework with learnable parameters. 
	
	\paragraph{The variational encoder} The variational encoder $p_{\theta}^{\rm enc}(z(t)\ |\ u(t)) := \mathcal{N}(m_z, S_z)$ involves two parametrized terms: a mean function that maps the QoI ${u}(t) \in \real{D}$ to the mean of the latent state  $m_z \in \real{d}$, and a variance function that maps the QoI ${u}(t) \in \real{D}$ to the variance of the latent state  $s_z \in \real{d}$, such that $S_z = \text{diag}(s_z)$. Note that in this definition, the covariance $S_z$ is parametrized as a diagonal matrix.
	
	\paragraph{The variational decoder} The variational decoder $p_{\theta}^{\rm dec}(u(t)\ |\ z(t)) := \mathcal{N}(m_u, S_u)$ also involves two parametrized terms: a mean function that maps the latent state ${z}(t) \in \real{d}$ to the mean of the QoI $m_u \in \real{D}$, and a variance function that maps the latent state ${z}(t) \in \real{d}$ to the variance of the QoI $s_u \in \real{d}$, such that $S_u = \text{diag}(s_u)$. As with the encoder, in this definition, the covariance $S_u$ is parametrized as a diagonal matrix.
	
	\paragraph{The SDE prior} The latent SDE prior, $\text{d}z = \psi_{\theta} (z, t; \mu, f(t)) \text{d}t + \Psi_{\theta}(t; \mu) \text{d}\beta$, also involves two parametrized terms: a drift function $\psi_{\theta} : \real{d} \times [0,T] \times \real{L} \times \real{F} \to \real{d}$, and a dispersion matrix $\Psi_{\theta}: [0,T] \times \real{L} \to \real{d}$. In all our numerical studies, we use feedforward neural networks with {\sf ReLU} activation to represent $\psi_\theta$ and $\Psi_\theta$. We apply the transformation $t \mapsto (\sin(2\pi t), \cos(2\pi t)) \in \mathbb{R}^2$ to the time variable $t$ and provide this feature vector as input to the feedforward neural network parametrizing the drift function. This feature space map is sometimes referred to as a ``positional encoding'' and widely used to overcome the spectral bias of coordinate-based feedforward neural networks; for a detailed discussion, see~\cite{mildenhall2020nerf,tancik2020fourfeat}.
	
	\paragraph{The amortized variational distribution} The amortized variational distribution, $q_{\phi}(z\ |\ t; \mu_i, f_i, \mathcal{D}_{i,k})$, is parametrized using a deep kernel model with learnable parameters $\sigma$, $\sigma_f$,  $\ell$, and the weights and biases of the feedforward neural network $\varphi$; see Section~\ref{sec:amortized}. This kernel maps from two time samples $t_1, t_2 \in \real{}$ to a similarity measure $\kappa(t_1, t_2) \in \real{}$.
	
	\section{Details of Example Problems}\label{sec:testdetails}
	
	This section provides additional details for the example problems given in Section~\ref{sec:examples}. For each case, we also describe the model architecture and the training settings. All the numerical studies were carried out on an Ubuntu server with a dual Intel Xeon E5-2680 v3 with a total of 24 cores, 128GB of RAM, and an Nvidia GeForce RTX 4090 GPU. PyTorch Lightning~\cite{Falcon_2019_PyTorch_Lightning} was used to implement the proposed stochastic ROM methodology, and the code along with test cases are available on our GitHub repository: \url{https://github.com/ailersic/visde}. In all the numerical studies, we use an amortized variational distribution for the latent state $z$. If prior distributions are specified for certain model parameters $\theta$, we learn variational distributions over those as well; if not, we obtain point estimates.
	
	We provide associated movies for each example problem as supplemetary material to this paper, showing model predictions and error.
	
	\begin{figure}[t!]
		\centering
		\includegraphics[width=11.4cm]{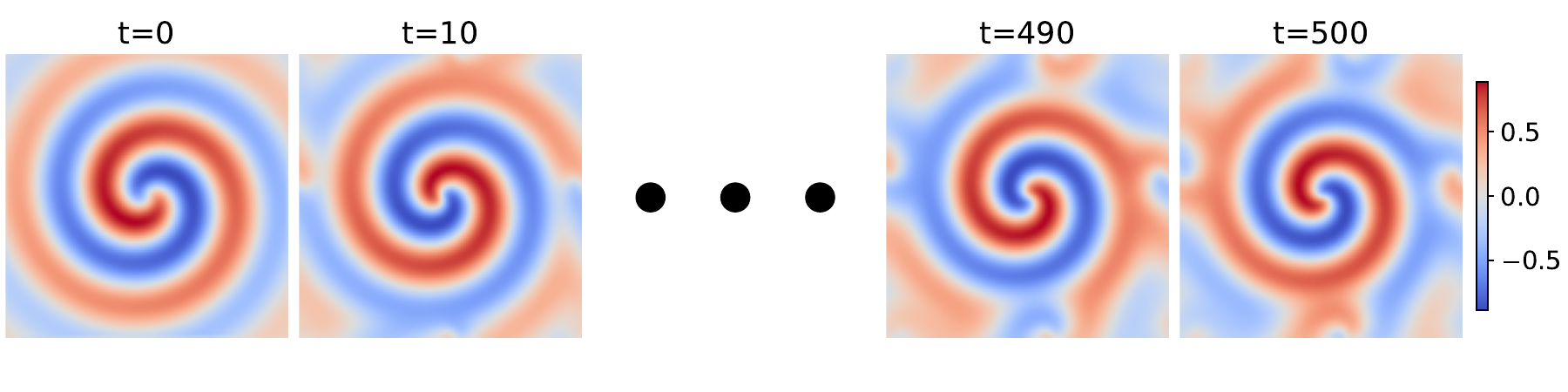}
		\caption{Reaction-diffusion: snapshots of solution over the whole time domain.} \label{fig:reactiondiffusionsol}
	\end{figure}
	
	\subsection{Reaction-Diffusion System in 2D} \label{sec:reacdiffdetails}
	The first example is the reaction-diffusion system taken from Champion et al.~\cite{Champion_2019_Data_Driven_Coordinates} and given by a coupled system of PDEs,
	\begin{equation*}
		\begin{aligned}
			v_{t} & = (1 - (v^2 + w^2))v + \beta (v^2 + w^2)w + \lambda (v_{xx} + v_{yy}) \\
			w_{t} & = (1 - (v^2 + w^2))w - \beta (v^2 + w^2)v + \lambda (w_{xx} + w_{yy}),
		\end{aligned}
	\end{equation*}
	with $(x, y, t) \in [-10, 10] \times [-10, 10] \times [0, 500]$, $\lambda = 0.1$, and $\beta=1$. The initial condition is given by
	\begin{equation*}
		\begin{aligned}
			v(x,y,0) & = \tanh\left(\sqrt{x^2 + y^2}\cos\left(\angle(x+iy) - \sqrt{x^2 + y^2}\right)\right) \\
			w(x,y,0) & = \tanh\left(\sqrt{x^2 + y^2}\sin\left(\angle(x+iy) - \sqrt{x^2 + y^2}\right)\right),
		\end{aligned}
	\end{equation*}
	where $i$ is the imaginary unit. The solution is shown in Figure~\ref{fig:reactiondiffusionsol}. This is an unparametrized and unforced test case, so $\mu$ and $f$ do not appear. Only $v$ is recorded in the dataset.
	
	The spatial domain is discretized using a grid with $D=100 \times 100$ equispaced points, and the time domain is discretized using 9999 steps. We consider a single trajectory, and the time interval is partitioned as follows: the interval $[0, 400]$ constitutes the training set, the next interval $(400, 450]$ is the validation set, and the final $(450, 500]$ is the test set. We set up the training data in a manner identical to Champion et al., except that whereas they corrupt the training data (both $v$ and its time derivative $v_t$, independently) using zero-mean Gaussian noise with standard deviation $10^{-6}$, we consider two cases: a noiseless case, and a noisy case with standard deviation $10^{-2}$.
	
	To enable direct comparison with Champion et al.'s approach (which combines an autoencoder with a sparse latent ODE), we represent our drift term $\psi_{\theta}$ as a third-order polynomial and use a $d=2$-dimensional latent state. We do not modify the SINDy model architecture and use their suggested training protocol ($3000$ epochs plus $1000$ refinement epochs). We train our stochastic ROM for $1000$ epochs.
	
	\begin{table}
		\centering
		\caption{Reaction-diffusion: hyperparameter combinations chosen for latent SINDy models to minimize validation error.}
		\label{tab:reactiondiffusion_sindyhyp}
		\begin{tabular}{r|cc|cc}
			\toprule
			& \multicolumn{2}{c}{POD-SINDy} & \multicolumn{2}{c}{AE-SINDy} \\
			Dataset & Poly. order & Threshold & Poly. order & Threshold \\
			\midrule
			Noiseless & $3$ & $0.1$ & $3$ & $0.03$ \\
			Noisy & $3$ & $0.1$ & $3$ & $0.1$ \\
			\bottomrule
		\end{tabular}
	\end{table}
	
	For the latent SINDy baselines, we have two primary hyperparameters to choose: the order of the polynomial basis, and the thresholding on the coefficients. To choose the threshold, we conduct a grid search over $\{0.001, 0.003, 0.01, 0.03, 0.1\}$ and choose the value that minimizes mean error $\varepsilon_\mu$ over the validation set. The polynomial order is chosen to be $3$ following Champion et al.~\cite{Champion_2019_Data_Driven_Coordinates}. Hyperparameters that led to unstable models on the validation set were disqualified. The chosen hyperparameters are shown in Table~\ref{tab:reactiondiffusion_sindyhyp}.
	
	\paragraph{Architecture and Training}
	
	The encoder mean is a CNN that takes ${u}$ as the input with the following architecture:
	\begin{equation*}
		\begin{aligned}
			& \textrm{Conv}[1, 8] \xrightarrow{\text{ReLU}} \textrm{Conv}[8, 16] \xrightarrow{\text{ReLU}} \textrm{Conv}[16, 32] \xrightarrow{\text{ReLU}} \\
			& \textrm{Conv}[32, 64] \xrightarrow{\text{ReLU}} \textrm{Conv}[64, 128] \xrightarrow{\text{ReLU}} \textrm{Linear}[n_h, d]
		\end{aligned}
	\end{equation*}
	where for all convolutional layers, stride is $2$, padding is $2$, and kernel size is $5$. Note that the first layer takes a $3$-channel input, as this dataset includes 2D velocity and pressure fields. The hidden dimension consequently becomes $n_h = 2048$ for this setting of the convolution layers. The architecture of the decoder mean  is defined in a symmetric manner using transposed convolution layers as follows:
	\begin{equation*}
		\begin{aligned}
			& \textrm{Linear}[d, n_h] \xrightarrow{\text{ReLU}} \textrm{ConvT}[128, 64] \xrightarrow{\text{ReLU}} \textrm{ConvT}[64, 32] \xrightarrow{\text{ReLU}} \\
			& \textrm{ConvT}[32, 16] \xrightarrow{\text{ReLU}} \textrm{ConvT}[16, 8] \xrightarrow{\text{ReLU}} \textrm{ConvT}[8, 1].
		\end{aligned}
	\end{equation*}
	Output padding is used to recover the state dimension $D$. The encoder variance and decoder variance are parametrized directly, such that they are constant with respect to $u$ and $z$, respectively.
	
	For this test case, the drift function is parametrized as a third-order polynomial of the latent state $z \in \real{d}$  with $d=2$, i.e., 
	\begin{equation*}
		\phi_{\theta}(z, t; \mu, f(t)) = \theta_1 + \theta_2 z_1 + \theta_3 z_2 + \theta_4 z_1^2 + \theta_5 z_1 z_2 + \theta_6 z_2^2 + \theta_7 z_1^3 + \theta_8 z_1^2 z_2 + \theta_9 z_1 z_2^2 + \theta_{10} z_2^3,
	\end{equation*}
	where $\theta_i \in \real{d}, i=1,2,\ldots,10$ leading to $\theta \in \real{10d}$ parameters in total. The parameters $\mu$ and forcing $f(t)$ are not relevant in this example, so they are not included in the drift. The dispersion matrix is, like the variance of the encoder/decoder, parametrized directly, making it constant with respect to $t$ and $\mu$.
	
	A log-normal prior is placed on the variance of the decoder, or equivalently, a normal prior is placed on the log-variance. This is to ensure that all components of the decoder variance stay within a similar order-of-magnitude range, which in turn reduces the variance in the gradient of the ELBO's log-likelihood term (see~\eqref{eqn:orig_elbo}).
	
	Finally, the deep kernel (see Section~\ref{sec:amortized}) applies the neural network $\varphi(t)$ to each input, which is defined as
	\begin{equation*}
		\textrm{Linear}[1, 128] \xrightarrow{\text{ReLU}} \textrm{Linear}[128, 128] \xrightarrow{\text{ReLU}} \textrm{Linear}[128, 1].
	\end{equation*}
	
	For this model, we train with the Adam optimizer~\cite{Kingma_2017_Adam} with an initial learning rate of $3\times 10^{-3}$ for $1000$ epochs and a batch size of $128$. We use an exponential decay scheduler, such that the learning rate decays by $10\%$ every $2000$ iterations.
	
	\begin{figure}[t!]
		\centering
		\includegraphics[width=11.4cm]{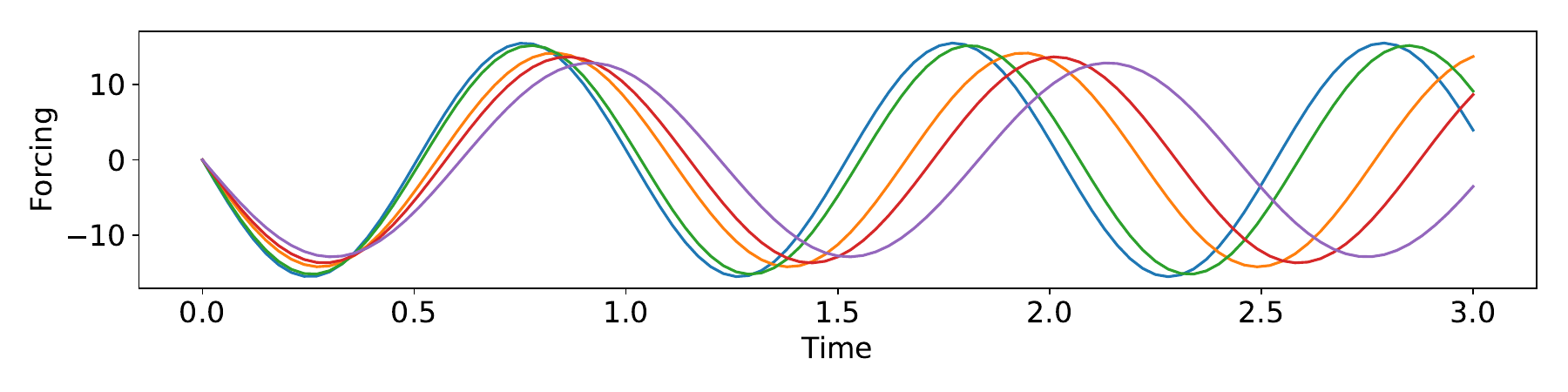}
		\caption{Burgers' equation: five samples of the forcing function $f$ in the training set.} \label{fig:burgersforcing}
	\end{figure}
	
	\begin{figure}[t!]
		\centering
		\includegraphics[width=11.4cm]{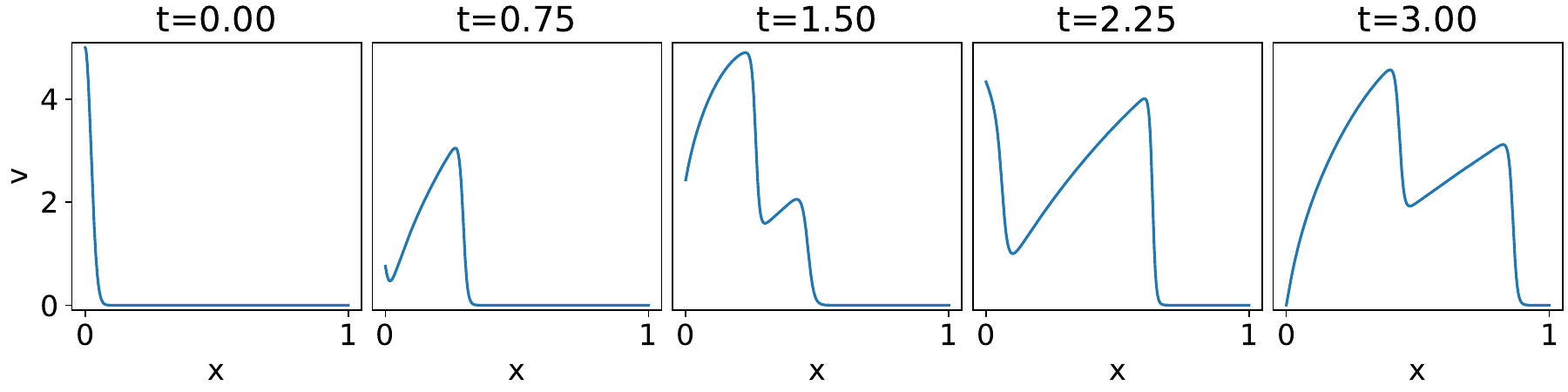}
		\caption{Burgers' equation: snapshots of solution over the whole time domain.} \label{fig:burgerssol}
	\end{figure}
	
	\paragraph{Interpretable dynamics}
	
	The strength of using SINDy for a problem like this is its ability to learn \emph{interpretable} dynamics. Champion et al.~\cite{Champion_2019_Data_Driven_Coordinates} reported the following linear latent system,
	\begin{equation*}
		\dot{z}_1 = -0.85 z_2 \qquad \dot{z}_2 = 0.97 z_1,
	\end{equation*}
	that lays bare the periodic nature of the dynamics. For each noisy/noiseless variant of the dataset, a similar linear system is discovered.\footnote{Note that the coefficient magnitudes are not unique since the autoencoder may allow the latent state to be arbitrarily scaled.} By using the same polynomial form for our drift function, we are able to recover similar linear dynamics. The drift function learned by our approach for the noiseless dataset is shown below:
	\begin{equation*}
		\begin{aligned}
			& \psi_{\theta}(z, t; \mu, f(t)) \\
			& = 10^{-2}\times \bmat{cccccccccc} \mbox{-}3.7 & 3.9 & \fbox{93} & \mbox{-}0.90 & \mbox{-}0.33 & \mbox{-}2.5 & \mbox{-}0.33 & 0.10 & \mbox{-}0.44 & 0.29 \\
			2.7 & \fbox{-89} & \mbox{-}1.9 & 0.37 & \mbox{-}1.0 & \mbox{-}0.48 & 0.24 & \mbox{-}0.50 & 0.33 & \mbox{-}0.51 \emat \bmat{c}1 \\ z_1 \\ z_2 \\ z_1^2 \\ z_1 z_2 \\ z_2^2 \\ z_1^3 \\ z_1^2 z_2 \\ z_1 z_2^2 \\ z_2^3 \emat \\ & \approx \bmat{c}0.93 z_2 \\ -0.89 z_1\emat.
		\end{aligned}
	\end{equation*}
	It can be noted that two of the coefficients are around $100$ times larger than the rest. This demonstrates that although our SVI-based approach does not require observations of the time-derivatives like SINDy, or any sparsity-inducing regularization or thresholding, it can yield similar sparse models. As mentioned in Section~\ref{sec:prior}, a sparsity-inducing prior~\cite{Course_2023_State_Estimation} can also be used in the proposed methodology to strongly enforce interpretability.
	
	\subsection{Forced, parametrized Burgers' equation} \label{sec:burgersdetails}
	The next case is the viscous one-dimensional Burgers' equation, given by
	\begin{equation*}
		v_t = -v\, v_x + \nu\, v_{xx},\quad (x,t) \in [0,1] \times [0,3],
	\end{equation*}
	with the boundary conditions $v(0,t) = f(t)$, $v(1,t) = 0$. The viscosity is parametrized as $\nu \in [0.05,0.1]$, and $f(t) = \alpha_1 \cos(2\pi\omega t)$ where $\omega \in [0.8, 1]$. The initial condition is
	$v(x, 0) = \alpha_1 \exp (-x^2/\alpha_2)$, where $\alpha_1 =5$ and $\alpha_2 = 0.001$.
	
	We discretize the governing equations on a spatial mesh with $D=500$ equispaced points. We evaluate the spatial derivatives with centered finite-difference operators and use a time marching scheme with $\Delta t = 0.001$ leading to a total of $1001$ time samples in each trajectory calculated using the FOM.	
	
	The training set consists of 100 trajectories computed for different realizations of $\mu = \nu$ and forcing function $f$, the validation set consists of 10, and the test set consists of 10. For each trajectory, we randomly sample the viscosity and forcing $\omega$ from uniform distributions, i.e., $\nu \sim \mathcal{U}[0.05, 0.1]$ and $\omega \sim \mathcal{U}[0.8, 1]$. Representative realizations of $f$ are shown in Figure~\ref{fig:burgersforcing} and a sample trajectory from the training set is shown in Figure~\ref{fig:burgerssol}. Because the PNODE and PNSDE models do not take the forcing function directly, for those models we include the frequency in the parameters, i.e., $\mu = \{\nu, \omega\}$.
	
	\begin{table}
		\centering
		\caption{Burgers' equation: hyperparameter combinations chosen for latent SINDy models to minimize validation error.}
		\label{tab:burgers_sindyhyp}
		\begin{tabular}{r|cc|cc}
			\toprule
			& \multicolumn{2}{c}{POD-SINDy} & \multicolumn{2}{c}{AE-SINDy} \\
			ROM dim. & Poly. order & Threshold & Poly. order & Threshold \\
			\midrule
			$d=3$ & $3$ & $0.003$ & $1$ & $0.1$ \\
			$d=4$ & $2$ & $0.03$ & $2$ & $0.1$ \\
			$d=5$ & $3$ & $0.03$ & $3$ & $0.1$ \\
			\bottomrule
		\end{tabular}
	\end{table}
	
	For the latent SINDy baseline, we have two primary hyperparameters to choose: the order of the polynomial basis, and the thresholding on the coefficients. To choose these, we conduct a grid search over polynomial orders $\{1, 2, 3\}$ and thresholds $\{0.001, 0.003, 0.01, 0.03, 0.1\}$ and choose the combination that minimizes mean error $\varepsilon_\mu$ over the validation set. Hyperparameters that led to unstable models on the validation set were disqualified. The chosen hyperparameters are shown in Table~\ref{tab:burgers_sindyhyp}.
	
	\paragraph{Architecture and training}
	
	The CNN architecture of the encoder mean that takes ${u}$ as the input is given below:
	\begin{equation*}
		\begin{aligned}
			& \textrm{Conv}[1, 8] \xrightarrow{\text{ReLU}} \textrm{Conv}[8, 16] \xrightarrow{\text{ReLU}} \textrm{Conv}[16, 32] \xrightarrow{\text{ReLU}} \\
			& \textrm{Conv}[32, 64] \xrightarrow{\text{ReLU}} \textrm{Conv}[64, 128] \xrightarrow{\text{ReLU}} \textrm{Linear}[n_h, d]
		\end{aligned}
	\end{equation*}
	where for all convolutional layers, stride is $2$, padding is $2$, and kernel size is $5$. The hidden dimension comes out to $n_h = 4096$ for this case. The architecture of the decoder mean that takes $z$ as the input is defined in a symmetric manner using transposed convolution layers as follows:
	\begin{equation*}
		\begin{aligned}
			& \textrm{Linear}[d, n_h] \xrightarrow{\text{ReLU}} \textrm{ConvT}[128, 64] \xrightarrow{\text{ReLU}} \textrm{ConvT}[64, 32] \xrightarrow{\text{ReLU}} \\
			& \textrm{ConvT}[32, 16] \xrightarrow{\text{ReLU}} \textrm{ConvT}[16, 8] \xrightarrow{\text{ReLU}} \textrm{ConvT}[8, 1]
		\end{aligned}
	\end{equation*}
	Output padding is used to recover the state dimension $D$. The encoder variance and decoder variance are parametrized directly, such that they are constant with respect to $u$ and $z$ respectively.
	
	As with the previous example, a log-normal prior is placed on the variance of the decoder. To show why this is necessary here, consider that the QoI $u(x, t) \approx 0$ near the right boundary of the $x$-domain for all $t$, $\mu$, and $f(t)$. Consequently, when the decoder mean function correctly predicts zero at the boundaries, decreasing the decoder variance causes the log-likelihood term to monotonically increase. Maximizing the ELBO therefore causes the decoder variance to tend to zero. Evaluating the ELBO may then lead to floating-point precision issues, or even a divide-by-zero error.
	
	The drift function is parametrized using a feedforward neural network with {\sf ReLU} activation that takes the latent state $z$, the viscosity $\nu$, forcing $f(t)$, and time as inputs,
	\begin{equation*}
		\textrm{Linear}[n_i, 128] \xrightarrow{\text{ReLU}} \textrm{Linear}[128, 128] \xrightarrow{\text{ReLU}} \textrm{Linear}[128, 128] \xrightarrow{\text{ReLU}} \textrm{Linear}[128, d]
	\end{equation*}
	where $n_i = d+1+N_f+2 = 13$. Note that the drift does not depend on the FOM initial condition parameters $\alpha_1$ and $\alpha_2$. The term 2 arises due to the feature space map we use for time as described earlier in Section~\ref{sec:modules}. The dispersion matrix is, like the variance of the encoder/decoder, parametrized directly, making it constant with respect to $t$ and $\mu$.
	
	Finally, the deep kernel (see Section~\ref{sec:amortized}) applies the neural network $\varphi(t)$ to each input, which is defined as
	\begin{equation*}
		\textrm{Linear}[1, 128] \xrightarrow{\text{ReLU}} \textrm{Linear}[128, 128] \xrightarrow{\text{ReLU}} \textrm{Linear}[128, 1].
	\end{equation*}
	
	For this model and all baseline models, we train using the Adam optimizer~\cite{Kingma_2017_Adam} with an initial learning rate of $10^{-3}$ for $50$ epochs and a batch size of $64$. We use an exponential decay scheduler, such that the learning rate decays by $10\%$ every $2000$ iterations.
	
	\subsection{Fluid flow with control input} \label{sec:flowdetails}
	
	The third example is adapted from the test case involving flow control of a two-dimensional Von K\'{a}rm\'{a}n vortex street at $Re=100$ studied by Rabault et al.~\cite{Rabault_2019_Flow_Control_DRL}. The normalized spatial domain is $x \in [-2, 2.1], y \in [-2, 20]$ and the normalized time domain is $t\in [0, 2]$, which covers $6\times$ the vortex shedding period $T_v \approx 0.33$. The time domain is discretized into $400$ time steps.
	
	The Navier-Stokes equations are solved on an unstructured mesh using FEniCS~\cite{Logg_2011_FEniCS} and the resulting two-dimensional velocity field and scalar pressure field are then resampled onto a $440\times 80$ uniform grid. 
	This gives a state dimension of $D = 440 \times 80 \times 3 = 105,600$, thereby enabling us to evaluate the scalability of our methodology.  
	The simulation's initial condition is a fully-developed flow with no forcing, so we allow the flow to develop with nonzero forcing for one period $T_v$ before including it in the dataset. Each trajectory, which now covers $t \in [0.33, 2]$ ($5\times T_v$), is then further partitioned into two trajectories covering $t \in [0.33, 1]$ and $t \in [1, 1.67]$ respectively. Each of these is then included in the dataset as a separate trajectory covering $2\times T_v$ with varying initial conditions. The time domain for each trajectory is $133$ time steps.

	\begin{figure}[t!]
		\centering
		\includegraphics[width=11.4cm]{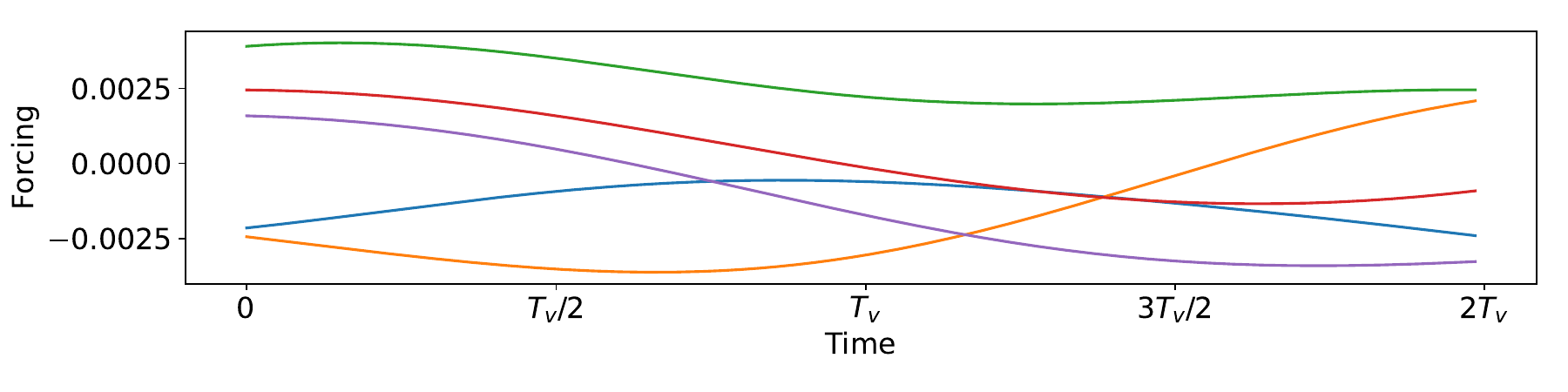}
		\caption{Fluid flow with control: five samples of the forcing function $f$ in the training set.} \label{fig:flowcontrolforcing}
	\end{figure}
	
	\begin{figure}[t!]
		\centering
		\includegraphics[width=\textwidth]{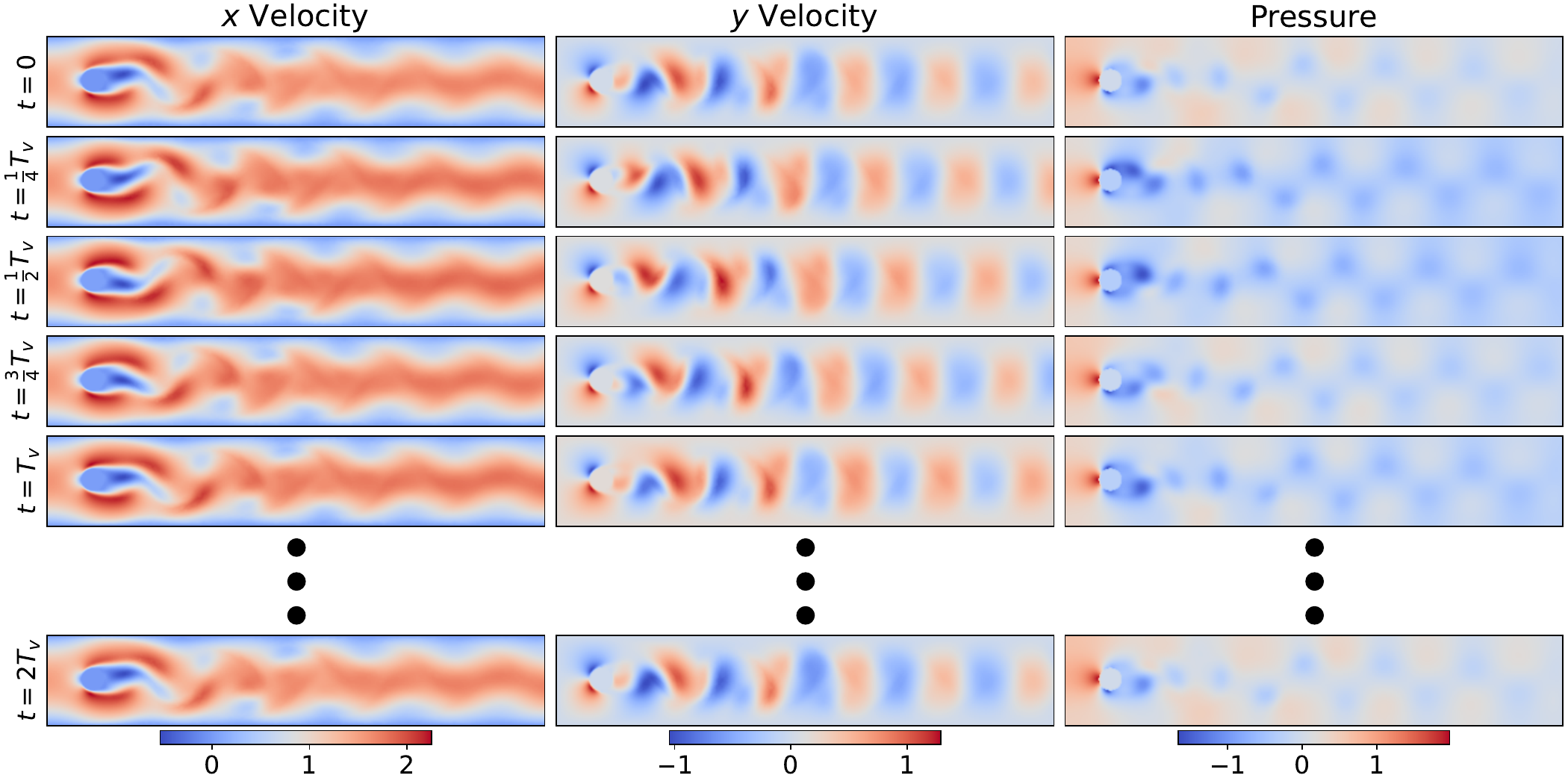}
		\caption{Fluid flow with control: snapshots of solution over the whole time domain.} \label{fig:flowcontrolsol}
	\end{figure}
	
	As discussed in Section~\ref{sec:examples}, the forcing $f$ corresponds to jets on the top and bottom of the cylinder which we parametrize as a zero-mean Gaussian process with covariance function $k(t_1, t_2) = \alpha \exp{(-(t_1 - t_2)^2/(2 \ell^2))}$, where $\ell = T_v$ and $\alpha = 10^{-5}$; this setting ensures that the flow rates are mostly within the interval $[-0.01,0.01]$. We generate 60 training, 5 validation, and 5 test trajectories for different realizations of $f$ sampled from $\mathcal{GP}(0,k)$. These are then partitioned as described above into 120 training, 10 validation, and 10 test trajectories. Representative samples of forcing are shown in Figure~\ref{fig:flowcontrolforcing}, and a sample trajectory from the training set is shown in Figure~\ref{fig:flowcontrolsol}.
	
	Because the forcing is not directly parametrized, it is not as straightforward as before to estimate the intrinsic solution manifold dimension. To approximate it, we collect the sampled forcings over $t \in [0, 2]$ and project them onto their POD basis. With 6 modes, we find we capture over 99\% of the kinetic energy. With time $t$ and 6 POD coefficients to represent forcing, we can uniquely specify any point in solution space, therefore 7 is taken to be our intrinsic solution manifold dimension, and our lower bound for the ROM dimension $d$.
	
	\begin{table}
		\centering
		\caption{Fluid flow with control: hyperparameter combinations chosen for latent SINDy models to minimize validation error.}
		\label{tab:flowcontrol_sindyhyp}
		\begin{tabular}{r|cc|cc}
			\toprule
			& \multicolumn{2}{c}{POD-SINDy} & \multicolumn{2}{c}{AE-SINDy} \\
			ROM dim. & Poly. order & Threshold & Poly. order & Threshold \\
			\midrule
			$d=7$ & $3$ & $0.001$ & $3$ & $0.001$ \\
			$d=8$ & $2$ & $0.1$ & $2$ & $0.1$ \\
			$d=9$ & $2$ & $0.03$ & $3$ & $0.001$ \\
			\bottomrule
		\end{tabular}
	\end{table}
	
	For the latent SINDy baseline, we have two primary hyperparameters to choose: the order of the polynomial basis, and the thresholding on the coefficients. To choose these, we conduct a grid search over polynomial orders $\{1, 2, 3\}$ and thresholds $\{0.001, 0.003, 0.01, 0.03, 0.1\}$ and choose the combination that minimizes mean error $\varepsilon_\mu$ over the validation set. Hyperparameters that led to unstable models on the validation set were disqualified. The chosen hyperparameters are shown in Table~\ref{tab:flowcontrol_sindyhyp}.
	
	\paragraph{Architecture and training}
	
	The encoder mean is a CNN that takes ${u}$ as the input with the following architecture:
	\begin{equation*}
		\begin{aligned}
			& \textrm{Conv}[3, 32] \xrightarrow{\text{ReLU}} \textrm{Conv}[32, 64] \xrightarrow{\text{ReLU}} \textrm{Conv}[64, 128] \xrightarrow{\text{ReLU}} \\
			& \textrm{Conv}[128, 256] \xrightarrow{\text{ReLU}} \textrm{Conv}[256, 512] \xrightarrow{\text{ReLU}} \textrm{Linear}[n_h, d]
		\end{aligned}
	\end{equation*}
	where for all convolutional layers, stride is $2$, padding is $2$, and kernel size is $5$. Note that the first layer takes a $3$-channel input since this dataset consists of two-dimensional velocity and pressure fields. The hidden dimension comes out to $n_h = 21504$ for this case. The architecture for the mean of the decoder that takes $z$ as the input is defined in a symmetric manner using transposed convolution layers as follows:
	\begin{equation*}
		\begin{aligned}
			& \textrm{Linear}[d, n_h] \xrightarrow{\text{ReLU}} \textrm{ConvT}[512, 256] \xrightarrow{\text{ReLU}} \textrm{ConvT}[256, 128] \xrightarrow{\text{ReLU}} \\
			& \textrm{ConvT}[128, 64] \xrightarrow{\text{ReLU}} \textrm{ConvT}[64, 32] \xrightarrow{\text{ReLU}} \textrm{ConvT}[32, 3]
		\end{aligned}
	\end{equation*}
	As in previous cases, output padding is used to recover the state dimension $D$. The encoder variance and decoder variance are parametrized directly, such that they are constant with respect to $u$ and $z$ respectively. Also as with the Burgers' equation test case, a log-normal prior is placed on the variance of the decoder.
	
	The drift function is a feedforward neural network that takes the latent state $z$, forcing $f(t)$, and time as inputs with the following architecture:
	\begin{equation*}
		\textrm{Linear}[n_i, 128] \xrightarrow{\text{ReLU}} \textrm{Linear}[128, 128] \xrightarrow{\text{ReLU}} \textrm{Linear}[128, 128] \xrightarrow{\text{ReLU}} \textrm{Linear}[128, d]
	\end{equation*}
	where $n_i = d+N_f= 10$. The parameters $\mu$ do not vary from trajectory to trajectory in this case, so it is not included in the drift. The dispersion matrix is, like the variance of the encoder/decoder, parametrized directly, making it constant with respect to $t$ and $\mu$.
	
	Finally, the deep kernel (see Section~\ref{sec:amortized}) applies the feedforward neural network $\varphi(t)$ to each input, which is defined as
	\begin{equation*}
		\textrm{Linear}[1, 128] \xrightarrow{\text{ReLU}} \textrm{Linear}[128, 512] \xrightarrow{\text{ReLU}} \textrm{Linear}[512, 128] \xrightarrow{\text{ReLU}} \textrm{Linear}[128, 1].
	\end{equation*}
	
	For this model and all baseline models, we train using the Adam optimizer~\cite{Kingma_2017_Adam} with an initial learning rate of $10^{-3}$ for $200$ epochs and a batch size of $64$. We use an exponential decay scheduler, such that the learning rate decays by $10\%$ every $2000$ iterations.

\end{document}